\documentclass[runningheads]{llncs}

\usepackage{eccv}

\usepackage{eccvabbrv}

\usepackage{graphicx}
\usepackage{booktabs}

\usepackage{algorithm}
\usepackage{algorithmic}

\usepackage{amsmath}
\usepackage{wrapfig}
\usepackage{adjustbox}
\usepackage{multirow}
\usepackage{minitoc}

\usepackage{amssymb}
\usepackage{mathtools}

\usepackage[accsupp]{axessibility}  

\usepackage{hyperref}

\usepackage{orcidlink}

\usepackage{xr-hyper}

\begin{document}

\title{LineGraph2Road: Structural Graph Reasoning on Line Graphs for Road Network Extraction} 

\titlerunning{LineGraph2Road}

\author{Zhengyang Wei\inst{1}\thanks{Corresponding author.}
\and
Renzhi Jing\inst{1}
\orcidlink{0000-0002-2116-1351} 
\and
Yiyi He\inst{2}
\orcidlink{0000-0002-9238-2880}
\and
Jenny Suckale\inst{1}
\orcidlink{0000-0002-9787-8180}
}

\authorrunning{Z.~Wei et al.}

\institute{Stanford University, Stanford CA 94305, USA \email{\{zywei,rjing,jsuckale\}@stanford.edu}\and
Georgia Institute of Technology, Atlanta GA 30332, USA\\
\email{yiyi.he@design.gatech.edu}}

\maketitle

\begin{abstract}
  Extracting routable road networks from satellite imagery requires accurate topology recovery beyond pixel-level segmentation. Recent methods decompose the task into keypoint detection and connectivity prediction, but reliably inferring road connectivity under structural ambiguity remains challenging. Overpasses introduce non-planar crossings that can create false shortcuts, and occlusions break visual continuity, requiring long-range structural reasoning. We formulate road extraction as fully unobserved connectivity inference and construct a global but sparse Euclidean graph from detected keypoints to better leverage the long-term context. To improve link prediction, we transform this graph into its line graph and perform reasoning using a Graph Transformer. We propose an end-to-end pipeline that integrates vision-based segmentation, sparse graph construction, and structured inference. Our method explicitly models overpasses and uses topology-preserving vertex extraction to mitigate routing-critical errors. This approach achieves state-of-the-art performance on City-scale, SpaceNet, and Global-scale benchmarks in topology metrics including TOPO-F1 and APLS. In addition to quantitative gains, our method improves reconstruction of multi-level road structures crucial for real-world routing reliability. The code is available at: \href{https://github.com/wzzzzzzy/LineGraph2Road}{https://github.com/wzzzzzzy/LineGraph2Road}.
\end{abstract}

\section{Introduction}
\label{sec:intro}

\begin{figure*}
  \centering
  \includegraphics[width=1\textwidth]{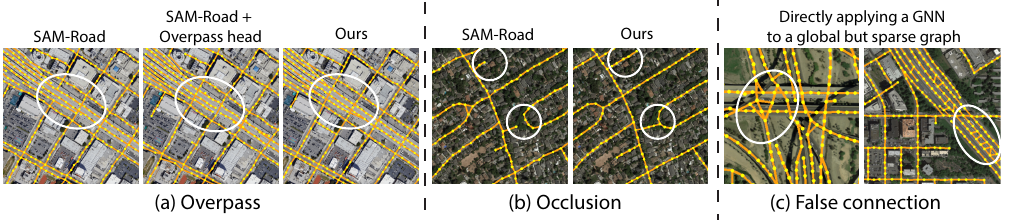}
  \caption{Challenges in road connectivity inference. 
    (a) Visual ambiguity at overpasses, where roads overlap visually but remain topologically disconnected, and overpass detection alone cannot fully resolve the connectivity uncertainty;
    (b) occlusions from trees or buildings that interrupt the visual continuity of roads;
    (c) naive GNN reasoning on sparse graphs that may still produce erroneous connections without appropriate structural modeling.}
  \label{fig:intro}
\end{figure*}

Road map extraction from satellite imagery is a popular and practically important task with broad applications in navigation~\cite{levinson2011towards} and urban management~\cite{chen2023semiroadexnet,mnih2010learning}. Early segmentation-based approaches~\cite{zhou2018d,bandara2022spin,batra2019improved,mattyus2017deeproadmapper,mosinska2018beyond} rely on heuristic post-processing~\cite{cheng2017automatic,mattyus2017deeproadmapper,costea2017creating} to recover topology, which can be error-prone. Iterative graph-growing methods~\cite{bastani2018roadtracer,tan2020vecroad,xu2022rngdet,xu2023rngdet++} alleviate some of these issues but suffer from limited parallelization. More recently, He \etal~\cite{he2022td} decomposed road extraction into keypoint prediction and connectivity prediction, a paradigm adopted by recent methods~\cite{hetang2024segment,yin2025towards}. However, incorrect connectivity predictions can still substantially distort routing distance estimation. One common challenge is occlusion, where roads are partially hidden by trees, buildings, or shadows, making continuous structures difficult to recover. Another major difficulty arises from overpasses~\cite{he2020sat2graph}, where non-planar crossing roads overlap in imagery and are not actually connected, often leading to erroneous topological predictions (\Cref{fig:intro}(a)).

We formulate road connectivity prediction as structured reasoning over a global but sparse Euclidean graph, rather than relying solely on local predictions~\cite{hetang2024segment} or fully dense pairwise modeling~\cite{shit2022relationformer,krzakala2024any2graph}. Specifically, we construct a graph by connecting candidate vertices within a predefined distance threshold, striking a balance between preserving global context and maintaining meaningful local structure. This design naturally incorporates long-range contextual information, which empirically helps resolve broken road segments caused by occlusion. However, applying Graph Neural Networks (GNNs)~\cite{kipf2016semi} to a global graph introduces challenges, including sensitivity to noisy candidate vertices and limitations in modeling edge representations, which can lead to an increased number of incorrect edges (\Cref{fig:intro}(c)), as also observed in prior work~\cite{bahl2022single}. 

To address this issue, we rely on prior work on line graphs~\cite{harary1960some,cai2021line,lachi2024simple} to transform the original graph into its line graph and perform connectivity inference as a node classification task using a Graph Transformer~\cite{shi2020masked}, rather than aggregating endpoint embeddings as in standard graph autoencoders~\cite{wang2019dynamic}. This representation improves structural expressiveness~\cite{zhang2021labeling} and enables more robust reasoning under occlusion through long-term context.

The overpass issue requires not only contextual learning but also accurate detection of crossing edge endpoints.
To do so, we augment the SAM-based encoder~\cite{kirillov2023segment,hetang2024segment,yin2025towards} with a dedicated head to explicitly predict overpass and underpass edge endpoints. Notably, purely local reasoning~\cite{hetang2024segment}, even with detected overpass endpoints, remains insufficient (\Cref{fig:intro}(a)), as overpasses often appear as partial occlusions and introducing additional candidate endpoints without global structural reasoning increases ambiguity and leads to more frequent misconnections. In contrast, combining contextual learning with accurate endpoint detection yields more reliable graph construction.

Our contributions can be summarized as follows:

\begin{itemize}
    \item We introduce an overpass/underpass segmentation head that explicitly differentiates multi-level crossings and a coupled NMS strategy that preserves critical connections around complex intersections. 

    \item We propose a paradigm where the road map extraction is formulated as binary classification over edges in a global but sparse Euclidean graph where only node pairs within predefined distance are considered rather than fully dense attention or local-only, make it possible to look into long-term context. 

    \item We integrate vision-based segmentation, sparse graph generation, and structured inference into an end-to-end pipeline, and empirically discovered that a line-graph transformation helps the GNN to learn efficient feature for effective road extraction.
    
    \item Extensive evaluation demonstrated that our method outperforms state-of-the-art methods under widely adopted metrics over three popular benchmarks. Additional ablation study and analysis are provided to verify the design choices and highlight the capability of our model in handling challenging cases including heavy occlusion and overpass.

\end{itemize}

\section{Related Work}

\subsection{Road Extraction from Satellite Images}
The methods for extracting road network graphs from high-resolution satellite images can be divided into two broad categories: segmentation-based and graph-based methods~\cite{he2020sat2graph}.

\textbf{Segmentation-based methods} aim to predict road masks using segmentation models, with various enhancements such as stronger backbones~\cite{zhou2018d,bandara2022spin}, joint orientation learning~\cite{batra2019improved}, and improved loss functions~\cite{mattyus2017deeproadmapper,mosinska2018beyond}. Some of these methods extract the road graph from the segmentation with post-processing heuristics, such as a thinning algorithm~\cite{cheng2017automatic}, A-star path-finding~\cite{mattyus2017deeproadmapper} or smoothing-based optimization algorithms~\cite{costea2017creating}. Because post-processing heuristics rely exclusively on predicted segmentation masks without incorporating image-level context, any noise or inaccuracies in the segmentation are inherently propagated to the extracted graph, limiting the accuracy of the topological connectivity inference~\cite{bastani2018roadtracer}.

\textbf{Graph-based methods} output the vectorized road network graph directly. RoadTracer~\cite{bastani2018roadtracer} pioneered an iterative search process guided by a CNN-based decision function, followed by improvements from VecRoad~\cite{tan2020vecroad}, RNGDet~\cite{xu2022rngdet}, and RNGDet++~\cite{xu2023rngdet++}, which introduced flexible step sizes, transformers with imitation learning, and instance segmentation heads, respectively. However, such vertex-by-vertex approaches are slow and hard to parallelize because of sequential dependencies, and errors can accumulate over time. 

In contrast, holistic methods generate the entire graph in a single forward pass~\cite{he2020sat2graph,he2022td,hetang2024segment,yin2025towards}. Sat2Graph~\cite{he2020sat2graph} encodes the graph as a tensor that captures keypoints and orientations. TD-Road~\cite{he2022td} employs two branches for keypoint prediction and relation reasoning. SAM-Road~\cite{hetang2024segment} builds on this with the Segment Anything encoder~\cite{kirillov2023segment} and a transformer-based reasoning module, while SAM-Road++~\cite{yin2025towards} enhances performance with node-guided resampling and extended-line strategies. kLCRNet~\cite{zhang2025klcrnet} uses a similar local connectivity module to directly construct connections between bipartite-matched keypoints using line pooling. However, the relational reasoning in these methods is restricted to each vertex and its immediate neighbors, overlooking potentially long-range dependencies beyond the predefined neighborhood.

\subsection{Graph Representation Learning}

Graph Neural Networks (GNNs)~\cite{kipf2016semi,velivckovic2017graph,wu2019simplifying} have wide applications in road attribute prediction~\cite{jepsen2019graph,he2020roadtagger,gharaee2021graph} and road representation learning~\cite{wu2020learning,zhou2024road}. Gharaee et al.~\cite{gharaee2021graph} employ a line-graph transformation for road type classification on a known road network, explicitly modeling edge–edge interactions and motivated by the limited features available to describe crossroads and intersections. For road network graph extraction, SPIN Road Mapper~\cite{bandara2022spin} performs graph reasoning on spatial and interaction spaces in the segmentation branch for a better segmentation mask, but does not apply directly to the construction of the road network graph. Bahl \etal~\cite{bahl2022single} use a GNN~\cite{wang2019dynamic} to predict links between the detected points of interest, but their approach did not outperform other methods. SAM-Road\cite{kirillov2023segment} designs its transformer as a form of Graph Convolutional Networks~\cite{kipf2016semi} as a topology decoder, but it only predicts in local subgraphs around individual vertices, overlooking the global topological structure and long-range dependency of the road network.

Image-to-graph methods like RelationFormer~\cite{shit2022relationformer} and Any2Graph~\cite{krzakala2024any2graph} tackle tasks such as network extraction~\cite{tetteh2020deepvesselnet} using transformer-based models which use full pairwise attention over all nodes, which is structurally uninformed. These methods typically operate on small, overlapping crops (\eg, 128 × 128 patches), limiting their ability to capture global context. Node identity issues arise because they lack mechanisms to maintain consistent node identifiers across patch boundaries. Reconstruction complexity further emerges as merging patch-level predictions into a single global graph requires non-trivial alignment and deduplication, often introducing additional errors and inconsistencies.

\section{Method}
\label{sec:method}
\subsection{Overall Architecture}
\begin{figure*}
  \centering
  \includegraphics[width=1\textwidth]{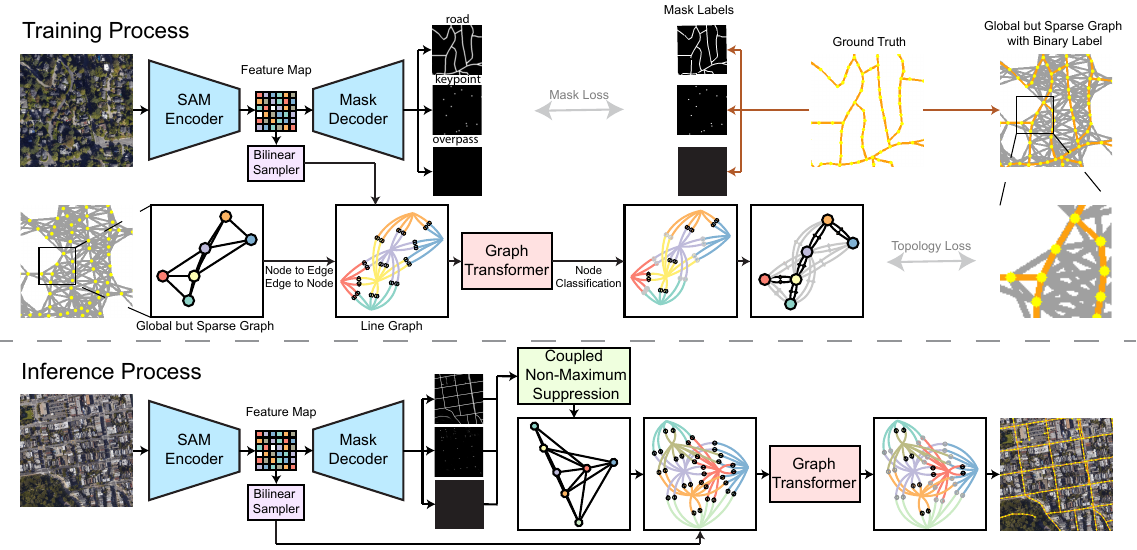}
  \caption{Overview of our LineGraph2Road pipeline. During both training and inference, a satellite image is processed by the pre-trained SAM encoder \cite{kirillov2023segment} to extract feature maps, which are decoded into road masks. In training, the graph is constructed from a preprocessed candidate edge set (brown lines), while in inference, it is built from vertices extracted via Coupled NMS on the road mask. This graph is converted into a line graph and passed to a Graph Transformer~\cite{shi2020masked} for binary node classification,  determining the presence of links and yielding the final road network graph.}
  \label{fig:framework}
\end{figure*}

\Cref{fig:framework} gives an overview of our LineGraph2Road pipeline. It consists of a pre-trained SAM~\cite{kirillov2023segment} image encoder and decoder for mask prediction (\Cref{sec:sam}), a coupled NMS module for vertex extraction (\Cref{sec:method_nms}), and a Graph Transformer-based module for connectedness prediction on the line graph (\Cref{sec:linegnn}). Taking an RGB satellite imagery as input, the SAM encoder generates feature maps and the decoder predicts probability maps for keypoints, roads, overpass/underpass crossings. Vertices are extracted from the probability maps by using coupled NMS. We treat vertex pairs within a distance threshold as candidate edges, with their features bilinearly sampled from the image feature map. These extracted vertices and candidate edges form a global but sparse Euclidean graph, which is then converted into its line graph. Finally, the line graph is processed by the Graph Transformer to predict binary labels for connectedness.

\subsection{Segment Anything Encoder and Decoder for Mask Prediction}
\label{sec:sam}
We use a pre-trained SAM~\cite{kirillov2023segment}, the ViT-B variant, as our image encoder. The encoder produces a feature map, $F \in \mathbb{R}^{\frac{H}{16}\times \frac{W}{16} \times D_{\text{feature}}}$, that is downsampled by a factor of 16 relative to the input image resolution $H \times W \times 3$. Following the original SAM design, we use the lightweight mask decoder without any input prompts. The decoder is configured for three output tasks, producing probability maps for the keypoint mask $\hat{M}_{\text{keypoint}}$, the road mask $\hat{M}_{\text{road}}$ and overpass/underpass mask $\hat{M}_{\text{overpass}}$. 

\subsection{Coupled Non-Maximum Suppression for Vertex Extraction}
\label{sec:method_nms}

Based on the keypoint, road and overpass/underpass masks, we adopt the idea of NMS~\cite{hosang2017learning} to obtain a set of sparse vertices which should be the node of the final road network graph.  We introduce a Coupled NMS algorithm instead of following the strategy~\cite{hetang2024segment}. In ~\cite{hetang2024segment}, vertices from two masks are separately extracted, and merged with priority given to intersection vertices, and suppressed by NMS again. In our Coupled NMS algorithm, vertices are first extracted from the keypoint and overpass/underpass mask using a keypoint suppression radius, $d_k$, and nearby points in the road mask are suppressed before extracting additional vertices from the road mask using a road suppression radius, $d_r$. The full procedure is detailed in \Cref{alg:cnms} (Appendix~\ref{sec:nms}). Our method not only reduces the failure cases described therein but also achieves a quantitative improvement in overall performance, as reported in \Cref{tab:nms}.

\subsection{Global but Sparse Euclidean Graph from Keypoints}

\begin{figure}
  \centering
  \includegraphics[width=1\textwidth]{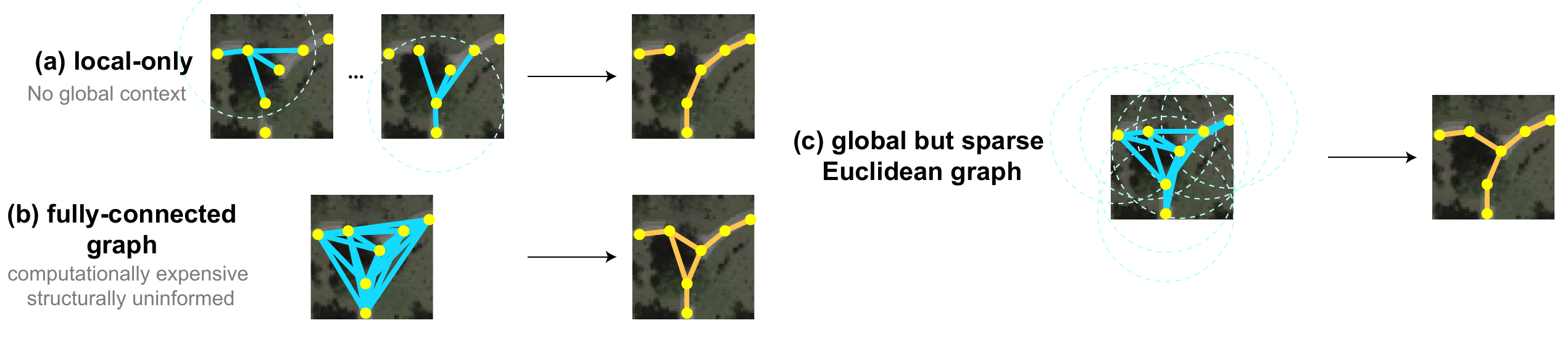}
  \caption{Comparison of graph connectivity strategies.
(a) Local-only (\eg, SAM-Road~\cite{hetang2024segment}) limits message passing to small subgraphs.
(b) Fully-connected (\eg, Any2Graph~\cite{krzakala2024any2graph}, RelationFormer~\cite{shit2022relationformer}) enables global reasoning but is computationally costly and structurally uninformed.
(c) global but sparse Euclidean graph connects nearby nodes within a distance threshold, balancing global context, structural awareness, and efficiency.}
  \label{fig:Euclidean}
\end{figure}

Coupled NMS yields a set of sparse vertices that are likely to be on the road network, but without connectedness information. To construct a graph that provides the model with broader contextual awareness of the road system, we treat vertex pairs within a distance threshold $d_{\text{nei}}$ as candidate edges. We label directly connected pairs as positive examples and others, including routable but indirect connections, as negative, creating a binary edge classification problem. The set of sparse vertices and the candidate edges can form a global but sparse Euclidean graph, $G = (V, E_{\text{nei}})$, where $E_{\text{nei}} = \left\{ (i, j) \mid i, j \in V,\ \left\| \text{pos}(i) - \text{pos}(j) \right\|_2 \leq d_{\text{nei}} \right\}$. This formulation instead connects node pairs within a distance threshold to form a global graph that balances global context and local structure, allowing the model to move beyond local heuristics, reduce computation, and infer connectivity in occluded or ambiguous regions as shown in \Cref{fig:Euclidean}. 

\subsection{Graph Transformer on Line Graph for Connectedness Prediction}
\label{sec:linegnn}

We obtain the feature for each candidate edge by sampling features from the feature map, $F$, using bilinear sampling at the two endpoints and a set of uniformly interpolated points. In total, we extract and concatenate $N_{\text{sampled}}$ features from the first endpoint to the second. This concatenated feature vector is then passed through a multilayer perceptron (MLP) to generate the candidate edge feature.

Beyond using visual features for each candidate edge, we also take into account how the edge fits into the overall structure of the graph. Rather than predicting edge presence by directly aggregating node representations~\cite{zhang2018link} of two endpoints of the edge, we first convert the original graph into a line graph~\cite{harary1960some}, $L(G)$.
\begin{definition}\textbf{(Line Graph)} The \emph{line graph} $L(G)$ of a graph $G$ is a graph where each node in $L(G)$ corresponds to an edge in $G$ and two nodes in $L(G)$ are adjacent if and only if their corresponding edges in $G$ share a common vertex.
\end{definition}

By converting the graph into its line graph, we reformulate the task as node classification, where each node in $L(G)$ represents a candidate edge of the original graph with its corresponding features. Prior work has explored line graphs for link prediction~\cite{cai2021line,lachi2024simple}. In our setting, this transformation enables the model to reason directly over edge representations rather than relying on endpoint feature aggregation, which can obscure structural differences between edges (see Appendix~\ref{sec:express}).

We employ a 3-layer Graph Transformer architecture~\cite{shi2020masked} to maintain a transformer-based design consistent with~\cite{hetang2024segment, yin2025towards}, ensuring that any performance gains arise from graph reasoning rather than differences in model class. For reference, we also compare our model against standard GNN baselines in Appendix~\ref{sec:express}.

Given candidate edge features derived from the image backbone, we apply multi-head self-attention over neighboring nodes in the line graph, followed by residual connections, feed-forward layers, and normalization. The final layer outputs a binary prediction for each candidate edge via a linear classifier with sigmoid activation. Additional implementation details of the 3-layer Graph Transformer are provided in Appendix~\ref{sec:graph_transformer}.

\subsection{Training Process}
\label{sec:train}
\normalsize

The training data used as model inputs or targets include: RGB satellite images, road masks, keypoint masks, overpass/underpass mask, and candidate edge pairs with their corresponding endpoint positions and binary labels that indicate whether two endpoints are directly connected.

To construct the road mask, ${M}_{\text{road}}$, we rasterize ground-truth road centerlines into a binary map following~\cite{hetang2024segment}. 
The keypoint mask, ${M}_{\text{keypoint}}$, is obtained by marking all vertices with degree $\neq 2$, as well as degree-2 vertices whose adjacent edges form angles between $60^\circ$ and $120^\circ$, rendered as circular regions with a radius of 3 pixels. 

For the overpass/underpass mask, $M_{\text{overpass}}$, we automatically derive supervision from the vectorized road graph. We first apply the graph refinement procedure in \Cref{alg:road_refine} (Appendix~\ref{sec:preprocess}) to separate geometrically overlapping but topologically disconnected crossings from true planar intersections. Edges that remain geometrically intersecting but are separated in the refined graph are identified as overpass/underpass crossings. We then rasterize the endpoints of these crossing edges as circular regions with a radius of 3 pixels, producing binary masks that supervise the overpass/underpass prediction head and guide Coupled NMS to preserve topology-critical vertices as shown in \Cref{fig:overpass_mask}.

We then derive candidate edge pairs from the vectorized road network graph. To align with the Coupled NMS vertex set, we include all graph keypoints and interpolate additional points along paths between them at random intervals in $[d_r, 2d_r]$, where $d_r$ is the NMS suppression radius (\Cref{alg:sampling_distances} in Appendix~\ref{sec:preprocess}). After applying Coupled NMS and adjusting the endpoints of overpass/underpass edges (\Cref{alg:road_refine} in Appendix~\ref{sec:preprocess}), candidate edges are formed by connecting vertex pairs within a distance threshold $d_{\text{nei}}$, with binary labels $B$ indicating whether the endpoints are directly connected in the ground-truth graph. The full preprocessing pipeline is illustrated in \Cref{fig:preprocess} in Appendix~\ref{sec:preprocess}.

The model is trained end-to-end using RGB images and candidate edge positions as input, and the segmentation masks and binary connectedness labels as supervision. For computational efficiency, images, masks, and graph structures are randomly cropped into smaller patches during training. Our loss function combines the binary cross-entropy (BCE) loss for both mask prediction and connectedness prediction:
\begin{equation}
    \mathcal{L}= \mathcal{L}_{BCE} (\hat{M}, {M})  + \lambda \cdot \mathcal{L}_{BCE} (\hat{B},B) 
\end{equation}
where $M = (M_{\text{keypoint}}, M_{\text{road}}, M_{\text{overpass}})$ and $\hat{M} = (\hat{M}_{\text{keypoint}}, \hat{M}_{\text{road}}, \hat{M}_{\text{overpass}})$.  

\subsection{Inference Process}
\label{sec:infer}
During inference, we pass satellite images through the SAM encoder and decoder to obtain the predicted segmentation masks, including the road mask, keypoint and overpass/underpass mask as shown in \Cref{fig:framework}. We follow the sliding-window approach and the threshold-selection approach in \cite{hetang2024segment} to tackle arbitrarily large regions. We first apply our coupled NMS to process the masks and extract a set of sparse vertices. We then identify candidate edges within a specified distance threshold, $d_{\text{nei}}$, construct a global but sparse Euclidean graph and convert to its line graph, and feed it into the Graph Transformer to predict the probability. Since the sliding-window approach may result in the same edge being predicted multiple times, we average the predicted probabilities to obtain the final confidence. 

\section{Experiments}

\subsection{Datasets}
\label{sec:dataset}

We evaluate our method on three benchmarks: City-scale~\cite{he2020sat2graph}, SpaceNet~\cite{van2018spacenet}, and Global-scale~\cite{yin2025towards}. City-scale contains 180 satellite images (2048×2048) from 20 U.S. cities, with 27 used for testing. SpaceNet includes 2,549 images (400×400) across Las Vegas, Paris, Shanghai, and Khartoum, with 382 for testing. Global-scale comprises 3,468 images (2048×2048) worldwide, with 624 for testing. 

All datasets have 1 m spatial resolution and provide ground-truth road graphs in adjacency-list format. We adopt the same train/val/test splits as~\cite{yin2025towards}.

\subsection{Metrics}

We look at two metrics, TOPO~\cite{he2018roadrunner} and APLS~\cite{van2018spacenet}.
TOPO~\cite{he2018roadrunner} evaluates the geometric and topological alignment between the ground truth and the proposal graphs. It places seed points along ground-truth graphs, samples subgraphs from seed points, and performs a one-to-one matching of points based on spatial and angular thresholds to compute precision, recall, and F1 scores.
APLS~\cite{van2018spacenet} measures the difference between the ground truth and proposal graphs with the differences in the shortest path lengths between corresponding nodes in both graphs.

\subsection{Implementation Details}
\label{sec:implementation}

For the City-scale and Global-scale dataset, we randomly crop 512$\times$512 patches from 2048$\times$ 2048 images with a batch size of 4. For the SpaceNet dataset, we use the original 400$\times$ 400 images with the same batch size. During training, we randomly rotate patches by multiples of 90 degrees to enhance robustness. The SAM encoder outputs feature maps with a dimensionality of $D_{\text{feature}}=128$. We set the keypoint suppression radius $d_k$ to 8 and the road suppression radius $d_r$ to 16. To construct the graph, vertex pairs within a local neighborhood threshold $d_{\text{nei}}=64$ are considered candidate edges. For each candidate edge, we extract $N_{\text{sampled}}=4$ positions of features from the feature map using bilinear interpolation along the edge. 

We set the number of attention heads for the Graph Transformer $C$ to 4. The dimension of the hidden layer in the Graph Transformer is 128. The dropout rate for the Graph Transformer is $0.1$. We use the Adam optimizer~\cite{kingma2014adam} with a base learning rate of 0.001 for all parameters, except for the pre-trained SAM encoder, which uses a learning rate scaled to $0.05 \times$ the base rate. The $\lambda$ in the loss function is set to 0.1. For the SpaceNet and City-scale datasets, we train LineGraph2Road for 10 epochs, consistent with the training process of SAM-Road~\cite{hetang2024segment} for fair comparison. For the Global-scale dataset, we train LineGraph2Road for 20 epochs, with a learning rate scheduler applied after 10 epochs using $\gamma = 0.1$. During inference, we apply a 16$\times$16 sliding window for the City-scale and Global-scale datasets, while the SpaceNet dataset is processed without sliding window. We conduct all experiments using four NVIDIA A100 GPUs.

\subsection{Comparative Results}

\begin{table*}
  \caption{Comparison with existing methods on City-scale and SpaceNet datasets.}
  \label{tab:comparison}
  \small
  \centering
  \begin{tabular}{l cccc cccc}
\toprule
 & \multicolumn{4}{c}{\textbf{City-scale Dataset}} & \multicolumn{4}{c}{\textbf{SpaceNet Dataset}}           \\ \cmidrule(r){2-5} \cmidrule(r){6-9}

{Methods}           & {Prec.↑} & {Rec.↑} & {F1↑} & {APLS↑} & {Prec.↑} & {Rec.↑} & {F1↑} & {APLS↑} \\ \hline

RoadTracer~\cite{bastani2018roadtracer}      & 78.00           & 57.44          & 66.16         & 57.29          & 78.61           & 62.45          & 69.90         & 56.03          \\ 
Sat2Graph~\cite{he2020sat2graph}      & 80.70           & 72.28          & 76.26         & 63.14          & 85.93           & 76.55          & 80.97         & 64.43          \\ 
TD-Road~\cite{he2022td}        & 81.94           & 71.63          & 76.43         & 65.74          & 84.81           & \textbf{77.80}          & 81.15         & 65.15          \\ 
RNGDet~\cite{xu2022rngdet}         & 85.97           & 69.78          & 76.87         & 65.75          & 90.91           & 73.25          & 81.13         & 65.61          \\ 
RNGDet++~\cite{xu2023rngdet++}       & 85.65           & 72.58          & 78.44         & 67.76          & 91.34           & 75.24          & 82.51         & 67.73          \\ 
SAM-Road  ~\cite{hetang2024segment}               & 90.47           & 67.69          & 77.23         & 68.37          & 93.03           & 70.97          & 80.52         & 71.64          \\ 
SAM-Road++ ~\cite{yin2025towards}                & 88.39       & 73.39 & 80.01 & 68.34        & \textbf{93.68} & 72.23    & 81.57         & 73.44       \\ 
kLCRNet ~\cite{zhang2025klcrnet}                & 86.77       & 71.59 & 78.27 & 68.96        & 89.69 & 72.20    & 80.00         & 67.51       \\ 

\textbf{Ours }                & {91.09}   &   {75.44}&{82.37}     &  {68.88}      & 92.82          & {77.11} & \textbf{84.24} & \textbf{73.94}          \\ 
\textbf{Ours w/ Overpass}                & \textbf{92.75}   &   \textbf{76.64}&\textbf{83.77}     &  \textbf{70.40}      & 93.50          & {76.38} & {84.08} & {73.36}          \\
\bottomrule
\end{tabular}
\end{table*}

We benchmark LineGraph2Road, with and without the overpass/underpass head, on City-scale and SpaceNet dataset against other methods as shown in \Cref{tab:comparison}. We compare several baselines, including RoadTracer~\cite{bastani2018roadtracer}, Sat2Graph~\cite{he2020sat2graph}, TD-Road~\cite{he2022td}, RNGDet~\cite{xu2022rngdet}, RNGDet++~\cite{xu2023rngdet++}, SAM-Road  ~\cite{hetang2024segment}, SAM-Road++ ~\cite{yin2025towards} and kLCRNet~\cite{zhang2025klcrnet}. 
On the City-scale dataset, LineGraph2Road with overpass head achieves a new state-of-the-art across all TOPO metrics, precision, recall, F1, and  APLS, with a clear improvement in topological accuracy. On the SpaceNet dataset, LineGraph2Road without overpass head achieves  the highest F1 and APLS. Although its precision (92.82) is slightly lower than the highest reported by SAM-Road++ (93.68), and its recall (77.11) is marginally below that of TD-Road (77.80), our method achieves the best balance between the two. Unlike previous methods that tend to favor high precision or high recall, LineGraph2Road maintains consistently strong performance on both, leading to a more complete and accurate road graph extraction than previous approaches. The stronger performance of the non-overpass variant on SpaceNet likely reflects the scarcity of overpass structures in this dataset (387 crossings over 408~km$^2$, compared to 4,180 over 720~km$^2$ in City-scale).

We also evaluate our model on the recently released Global-scale dataset, using in-domain test sets, as summarized in \Cref{tab:comparison_global_indomain}. LineGraph2Road achieves the highest scores in recall, F1, and APLS, although its precision is relatively lower. This reflects a recall-oriented trade-off: our method recovers more valid long-range connections, which improves network completeness but can introduce a small number of extra edges.
The improvement is most notable on APLS, where LineGraph2Road reaches 68.70, substantially higher than the previous best of 62.19, demonstrating its strong ability to capture road connectivity. The failure cases of Global-scale dataset are further analyzed in the Appendix~\ref{sec:failureglobal}.

\begin{table}[ht]
  \caption{Comparison with existing methods on Global-Scale (In-Domain) dataset.}
  \label{tab:comparison_global_indomain}
  \small
  \centering
  \begin{tabular}{l cccc}
\toprule
 & \multicolumn{4}{c}{\textbf{Global-Scale (In-Domain)}} \\ \cmidrule(r){2-5}

{Methods}           & {Prec.↑} & {Rec.↑} & {F1↑} & {APLS↑} \\ \hline

Sat2Graph~\cite{he2020sat2graph}      & 90.15  & 22.13 & 35.53 & 26.77 \\ 
RNGDet~\cite{xu2022rngdet}            & 79.89  & 40.72 & 52.59 & 49.43 \\ 
RNGDet++~\cite{xu2023rngdet++}        & 79.02  & 45.23 & 55.04 & 52.72 \\ 
SAM-Road~\cite{hetang2024segment}     & \textbf{91.93} & 45.64 & 59.80 & 59.08 \\ 
SAM-Road++~\cite{yin2025towards}      & 88.95  & 49.27 & 62.33 & 62.19 \\ 
\textbf{Ours }                         & 90.54  & \textbf{51.12} & \textbf{64.43} & 67.69 \\ 
\textbf{Ours w/ Overpass}              & 89.65  & 50.85 & 64.05 & \textbf{68.70} \\ 

\bottomrule
\end{tabular}
\end{table}

\begin{figure}[ht]
  \centering
  \includegraphics[width=1\textwidth]{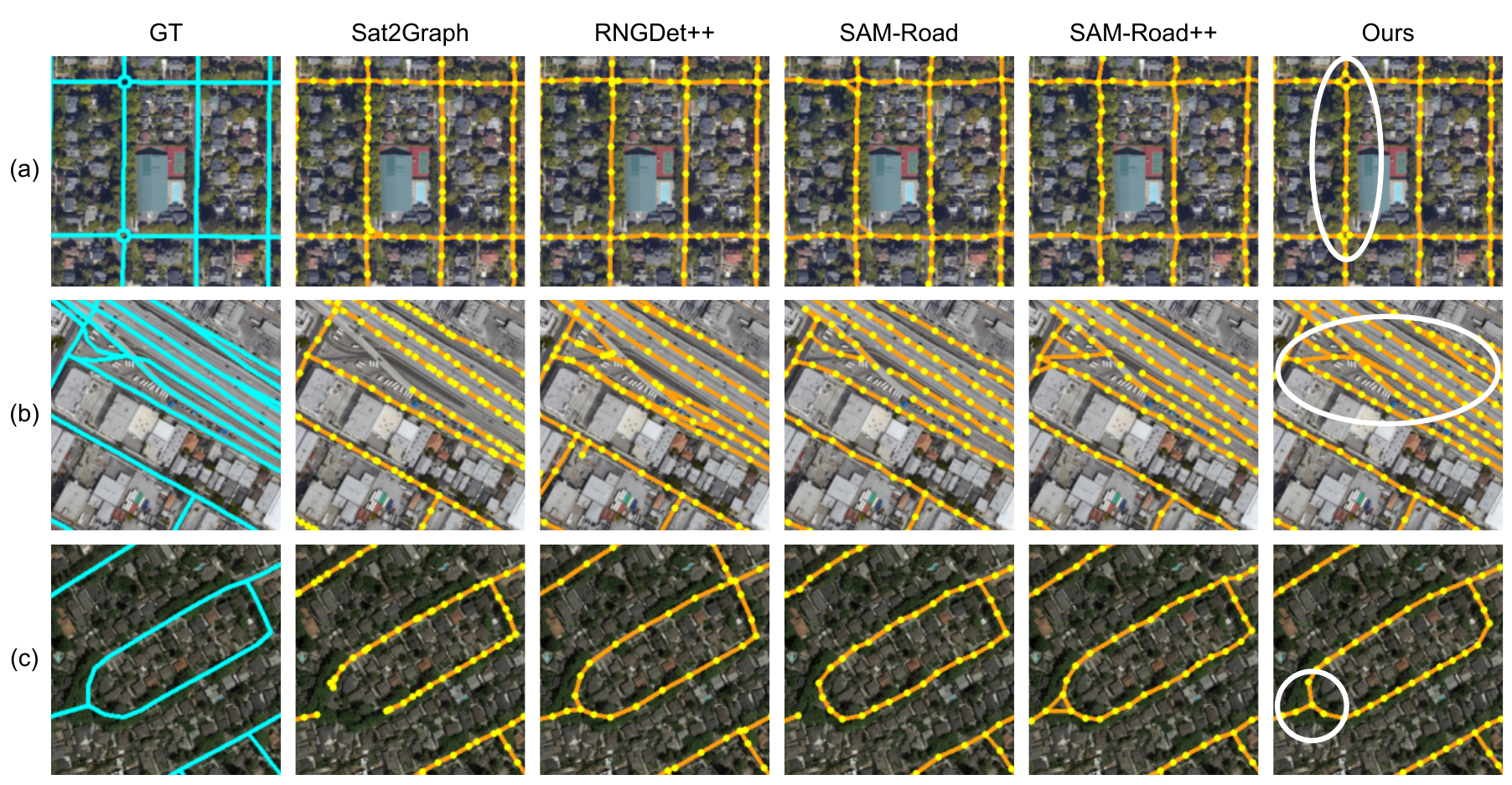}
  \caption{The visualized road network graph predictions of our method and other baseline methods on City-Scale dataset. (a) small roundabouts; (b) multi-lane highways with complex merges and curves; and (c) winding roads with irregular geometries.}
  \label{fig:visualized}
\end{figure}

Qualitative examples of road networks predicted by LineGraph2Road are shown in \Cref{fig:visualized} and \Cref{fig:vizoverpass}. The visualizations are shown alongside four competitive baselines~\cite{he2020sat2graph,xu2023rngdet++,hetang2024segment,yin2025towards} and the corresponding ground truth annotations. In general, LineGraph2Road shows strong performance in challenging conditions, such as small roundabouts at intersections, multilane highways with complex merges and curves, and winding roads with irregular geometries as shown in \Cref{fig:visualized}.

\subsection{Ablation Studies}

\begin{table*}
  \caption{Ablation study on City-scale dataset.}
  \label{tab:ablation}
  \small
  \centering
  \begin{adjustbox}{max width=0.95\textwidth}
  \begin{tabular}{c ccccc cccc}
    \toprule
         & \multicolumn{5}{c}{\textbf{Setting}}  & \multicolumn{4}{c}{\textbf{City-scale Dataset}} \\
        {Variant} & {SAM} & {Image Feature} & {Line Graph} & {Graph Transformer} & {$N_{\text{sampled}}$} & {Prec.↑} & {Rec.↑} & {F1↑} & {APLS↑} \\
        \midrule
--      & \checkmark & \checkmark & \checkmark & \checkmark & 4 & 91.09 & 75.44 & 82.37 & 68.88 \\
A      &            & \checkmark & \checkmark & \checkmark & 4 & 89.29 & 68.19 & 77.08 & 61.63 \\
B      & \checkmark &            & \checkmark & \checkmark & 4 & 80.93 & 72.20 & 76.12 & 69.17 \\
C      & \checkmark & \checkmark & \checkmark &            & 4 & 92.12 & 65.51 & 76.12 & 55.84 \\
D      & \checkmark & \checkmark & \checkmark & {to 3-layer MLP} & 4 & 93.03 & 62.36 & 74.20 & 52.20 \\
E      & \checkmark & \checkmark & \checkmark & \checkmark & 8 & 91.29 & 75.27 & 82.34 & 68.91 \\
F      & \checkmark & \checkmark & \checkmark & \checkmark & 2 & 91.29 & 75.11 & 82.25 & 68.11 \\
G      & \checkmark & \checkmark &  & \checkmark & 2 & 91.81 & 73.34 & 81.35 & 65.72 \\

H      & \checkmark & \checkmark &  & to RelationFormer decoder~\cite{shit2022relationformer} & 2 & 92.67 & 69.07 & 78.82 & 60.70 \\
I      & \checkmark & \checkmark &  & to Any2Graph decoder~\cite{krzakala2024any2graph} & 2 & 92.98 & 66.96 & 77.53 & 57.38 \\
\bottomrule
\end{tabular}
\end{adjustbox}
\end{table*}

We conduct ablation experiments to evaluate the impact of key components in LineGraph2Road on the City-scale dataset. To ensure fair comparison, we use the non-overpass version to eliminate variability introduced by the overpass head as shown in Appendix~\ref{sec:overpass_effect}. Results are shown in \Cref{tab:ablation}.

\textbf{Pretrained SAM.} For Variant A, we use the ViT-B encoder without pretrained weights from SAM. Under identical training conditions, the variant with pretrained weights significantly outperforms the one without, demonstrating the effectiveness of pretrained initialization.

\textbf{Visual features for guiding connectedness prediction.} In Variant B, we replace SAM-derived image features with a learnable pseudo-feature map for candidate edge embedding to isolate the role of visual cues. Although vertex extraction still uses SAM, connectivity prediction relies solely on the pseudo-features. This results in a substantial drop in precision and F1, indicating that visual features are essential for rejecting visually implausible yet topologically valid connections. Location cues alone are insufficient for reliable inference. Notably, APLS increases slightly, suggesting that it may reward topological continuity even when incorrect visual connections are introduced.

\textbf{Graph Transformer.} In Variant C, we remove the Graph Transformer and predict links independently using a fully connected layer, eliminating graph context. Both TOPO and APLS drop significantly. In Variant D, we replace the Graph Transformer with a 3-layer MLP, removing message passing via attention, which further degrades performance. These results demonstrate that relational reasoning is essential—visual features alone are insufficient for accurate connectivity inference without structural context.



\textbf{Number of sampled features per candidate edge.} For Variants E and F, we modify the number of sampled features per candidate edge to 8 and 2, respectively, and compare the results with the default setting of 4. The results show minimal variation in both TOPO and APLS metrics, suggesting that increasing the sampling density along a candidate edge provides limited additional information. This is expected for two reasons. First, the candidate graph is already local in Euclidean space because candidate edges are restricted by $d_{\text{nei}}$, so most edges cover relatively short image regions. Second, the SAM feature map is downsampled by a factor of 16, meaning that additional samples along the same short edge are often spatially close in feature space and provide redundant visual cues. Therefore, the main performance gain does not come from denser interpolation along individual edges, but from representing candidate edges as nodes in the line graph and allowing the Graph Transformer to reason over their structural relationships.

\textbf{Superiority of the line graph transformation over the original graph.} For Variant G, we conduct an ablation comparing the Graph Transformer applied on the original graph versus on the line graph, while keeping all other components, including SAM encoder, mask decoder, coupled NMS, bilinear sampler, identical. We used the sampled feature as node features as the original graph and applied a Graph Transformer with the same architecture on the original graph, and concatenated the output feature of two endpoints for each edge and passed it to a 3-layer MLP to get the final prediction and compare it with our line graph version.
Compared to Variant F with identical features, the original graph variant underperforms significantly, particularly on the APLS metric (65.71 vs 68.11), supporting our analysis in Appendix~\ref{sec:express} that aggregating node representations in a standard GNN constrains the expressiveness of structural link representations. We provide additional comparisons with alternative original-graph readouts in Appendix~\ref{sec:app_readout}, where the line-graph formulation remains superior.

\textbf{Comparison with other link prediction modules.} For Variants H and I, we keep the SAM encoder–decoder fixed and replace only the link prediction modules with those from RelationFormer~\cite{shit2022relationformer} and Any2Graph~\cite{krzakala2024any2graph}. This setup allows us to isolate and verify that the performance improvements originate from our Graph Transformer on the Line Graph, rather than from the powerful SAM backbone. When compared to Variant G, their performance is even worse, likely because the Graph Transformer in Variant G operates on a global but sparse Euclidean graph, whereas RelationFormer and Any2Graph perform on a fully connected graph that lacks structural priors.

\subsection{Comparative Analysis of Coupled vs. Standard NMS}
\label{sec:nmsvs}

To evaluate the effectiveness of our proposed Coupled NMS, we compare it with the standard NMS used in prior work~\cite{hetang2024segment} for our proposed method with all other components fixed on both the City-scale and SpaceNet datasets. Results are summarized in \Cref{tab:nms}. We observe that Coupled NMS consistently improves the performance of our method, yielding higher recall, TOPO-F1, and APLS on both datasets. The improvement is particularly clear when the overpass/underpass head is enabled, where preserving vertices near crossing structures avoids suppressing topology-critical endpoints.

\begin{table*}[ht]
  
  \caption{Performance comparison of NMS and Coupled NMS on the City-scale and SpaceNet datasets.}
  \label{tab:nms}
  \centering
  \begin{adjustbox}{max width=1\textwidth}
  \begin{tabular}{l cccc cccc}
\toprule
 & \multicolumn{4}{c}{\textbf{City-scale Dataset}} & \multicolumn{4}{c}{\textbf{SpaceNet Dataset}}           \\ \cmidrule(r){2-5} \cmidrule(r){6-9}

{ Methods}           & {\ Prec.↑} & {Rec.↑} & { F1↑} & {\ APLS↑} & { Prec.↑} & { Rec.↑} & { F1↑} & { APLS↑} \\ \hline

{\textbf{Ours} w/ NMS }   & \textbf{91.62}           & 74.65          & 82.11         & 68.57          & \textbf{93.47}           & 75.38          & 83.46         & 73.05 \\ 
{\textbf{Ours}   w/ Coupled}               & {91.09}   &   \textbf{75.44}&\textbf{82.37}     &  \textbf{68.88}      & 92.82          & \textbf{77.11} & \textbf{84.24} & \textbf{73.94}            \\ 

{\textbf{Ours} w/ Overpass w/ NMS }   & {92.89}           & 73.87         & 82.11         & 68.85          & {93.29}           & 75.13          & 83.23        & 72.63 \\ 
{\textbf{Ours} w/ Overpass w/ Coupled}                & \textbf{92.75}   &   \textbf{76.64}&\textbf{83.77}     &  \textbf{70.40}      & \textbf{93.50}          & \textbf{76.38} & \textbf{84.08} & \textbf{73.36}             \\ 
\bottomrule

\end{tabular}
\end{adjustbox}

\raggedright
\end{table*}

\subsection{Detection of Overpasses}

We evaluate overpass detection by identifying non-planar crossing road segments in ground-truth and predicted graphs. We convert each crossing into a bounding box enclosing the two segments. We then match predicted and ground-truth overpasses one-to-one using an IoU threshold of 0.25. We count matched pairs as true positives (TP), unmatched ground truth as false negatives (FN), and unmatched predictions as false positives (FP). Precision, Recall, and F1-score are computed per tile and aggregated across the dataset using summed TP, FP, and FN.

\begin{table}[ht]
\centering
\caption{Comparison of Overpass Detection Performance}
\label{tab:overpas_results}
\begin{tabular}{l c c c c c c}
\toprule
Method & TP & FP & FN & Prec. & Rec. & F1 \\
\hline
SAM-Road  ~\cite{hetang2024segment}    & 296 & 216 & 1734 & 57.81 & 14.58 & 23.29 \\

SAM-Road  ~\cite{hetang2024segment}  w/ Overpass     & 413 & 261 & 1617 & 61.28 & 20.34 & 30.55 \\
\textbf{Ours }    & 440 & 200 & 1590 & 68.75 & 21.67 & 32.96 \\

\textbf{Ours w/ Overpass}       & 740 & 208 & 1290 & \textbf{78.06} & \textbf{36.45} & \textbf{49.70} \\

\bottomrule
\end{tabular}
\end{table}

As shown in Table \ref{tab:overpas_results}, our method consistently outperforms the SAM-based baseline in both overpass and non-overpass variants. In particular, our model with overpass modeling achieves the highest F1-score of 49.70, substantially improving recall compared to all other variants. Although precision is relatively comparable across methods, the main performance gap arises from recall. These results indicate that integrating graph context with the dedicated overpass head is critical for accurately detecting stacking road structures and achieving overall superior performance.

\begin{figure*}
  \centering
  \includegraphics[width=1\textwidth]{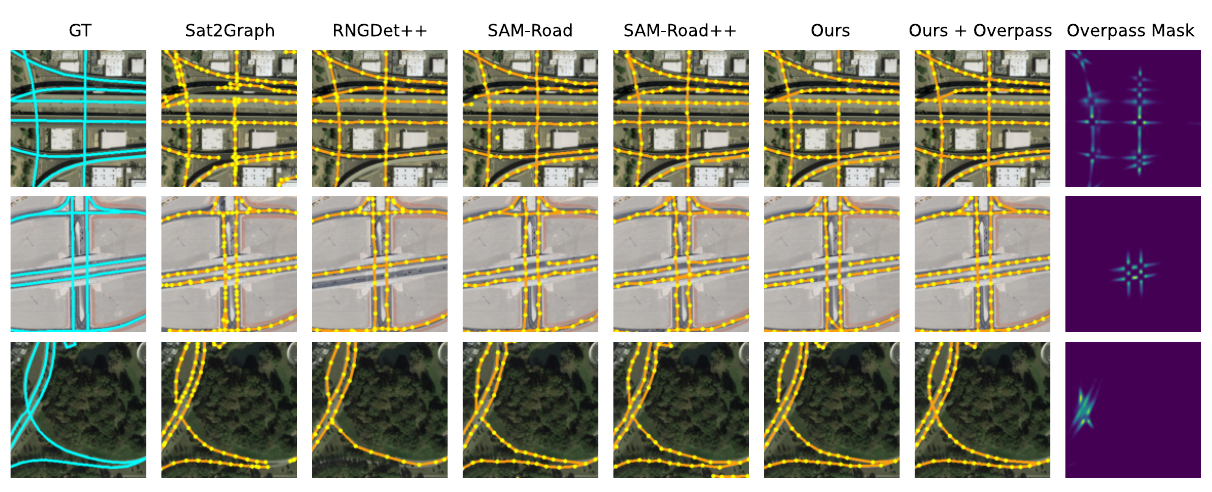}
  \caption{The visualized road network graph predictions of our method and other baseline methods as well as the predicted overpass/underpass mask on City-Scale dataset, illustrating performance on overpass and underpass crossings.}
  \label{fig:vizoverpass}
\end{figure*}

\Cref{fig:vizoverpass} presents overpass/underpass cases comparing predictions from baseline methods and our non-overpass variant, along with the corresponding overpass/underpass mask. The results illustrate that our model effectively identifies the relevant regions and reliably reconstructs multi-level crossings, whereas other methods fail to distinguish or maintain their connectivity.
We provide additional qualitative examples in the Appendix~\ref{sec:addviscity}.

\section{Conclusion}
\label{sec:cbl}
We present LineGraph2Road, an end-to-end framework for extracting road network graphs from satellite images that formulates the task as connectivity inference over candidate edges on a global but sparse Euclidean graph. This formulation enables topology recovery directly from detected vertices without assuming a known graph structure. To reason over ambiguous connections, we perform connectivity inference on the line graph of this graph using a Graph Transformer. Combined with explicit overpass/underpass modeling and a coupled NMS strategy for topology-preserving vertex extraction, the proposed system addresses common challenges such as non-planar crossings and occlusions. Our experiments demonstrate that LineGraph2Road achieves better performance on three existing public datasets without introducing significant inference overhead. Our framework advances our ability to automatically detect accurate road maps, including complex non-planar structures such as highway overpasses, which can benefit navigation, disaster response, and urban planning, particularly in regions lacking detailed infrastructure maps or after events that destroyed significant portions of the road infrastructure.

\section*{Acknowledgements}
We gratefully acknowledge support from the Stanford Sustainability Accelerator.

\bibliographystyle{splncs04}
\bibliography{main}

\end{document}


\appendix

\section{Coupled Non-Maximum Suppression}
\label{sec:nms}

\subsection{Algorithm}
 Instead of following the original strategy~\cite{hetang2024segment} that separately extracts vertices from both masks, merges them with priority given to intersection vertices, and then applies NMS again, we introduce a Coupled NMS algorithm. In this approach, vertices are first extracted from the keypoint mask using a keypoint suppression radius $d_k$, and nearby points in the road mask are suppressed before extracting additional vertices from the road mask using a road suppression radius $d_r$. 

\begin{algorithm}
\caption{Coupled Non-Maximum Suppression of Vertices}
\label{alg:cnms}
\begin{algorithmic}[1]
\STATE $V, V_k, V_o \leftarrow \emptyset$
\STATE $t_k, t_r \leftarrow$ keypoint and road threshold values
\STATE $d_k, d_r \leftarrow$ suppression radius for keypoint and road

\FOR{each pixel $(x, y)$ in the keypoint mask}
    \IF{pixel value $> t_k$}
        \STATE Add $(x, y)$ to $V_o$
    \ENDIF
\ENDFOR
\STATE Sort $V_o$ by pixel values in descending order
\FOR{each $(x, y)$ in $V_o$}
    \STATE Remove all $(x', y')$ after $(x, y)$ in $V_o$ where $\text{distance}((x', y'), (x, y)) < d_k$
\ENDFOR
\vspace{0.5em}
\FOR{each pixel $(x, y)$ in the overpass/underpass mask}
    \IF{pixel value $> t_k$}
        \STATE Add $(x, y)$ to $V_k$
    \ENDIF
\ENDFOR
\STATE Sort $V_k$ by pixel values in descending order
\FOR{each $(x, y)$ in $V_k$}
    \STATE Remove all $(x', y')$ after $(x, y)$ in $V_k$ where $\text{distance}((x', y'), (x, y)) < d_k$
\ENDFOR

\STATE$V_k \leftarrow V_k \cup V_o$
\FOR{each pixel $(x, y)$ in the road mask}
    \IF{pixel value $> t_r$}
        \STATE Add $(x, y)$ to $V$
    \ENDIF
\ENDFOR
\FOR{each $(x_k, y_k)$ in $V_k$}
    \STATE Remove all $(x, y)$ from $V$ where $\text{distance}((x, y), (x_k, y_k)) < d_r$
\ENDFOR
\STATE Sort $V$ by pixel values in descending order
\FOR{each $(x, y)$ in $V$}
    \STATE Remove all $(x', y')$ after $(x, y)$ in $V$ where $\text{distance}((x', y'), (x, y)) < d_r$
\ENDFOR
\STATE $V \leftarrow V \cup V_k$
\RETURN $V$
\end{algorithmic}
\end{algorithm}

\subsection{Analysis}
\begin{figure*}[ht]
  \centering
  \includegraphics[width=1\textwidth]{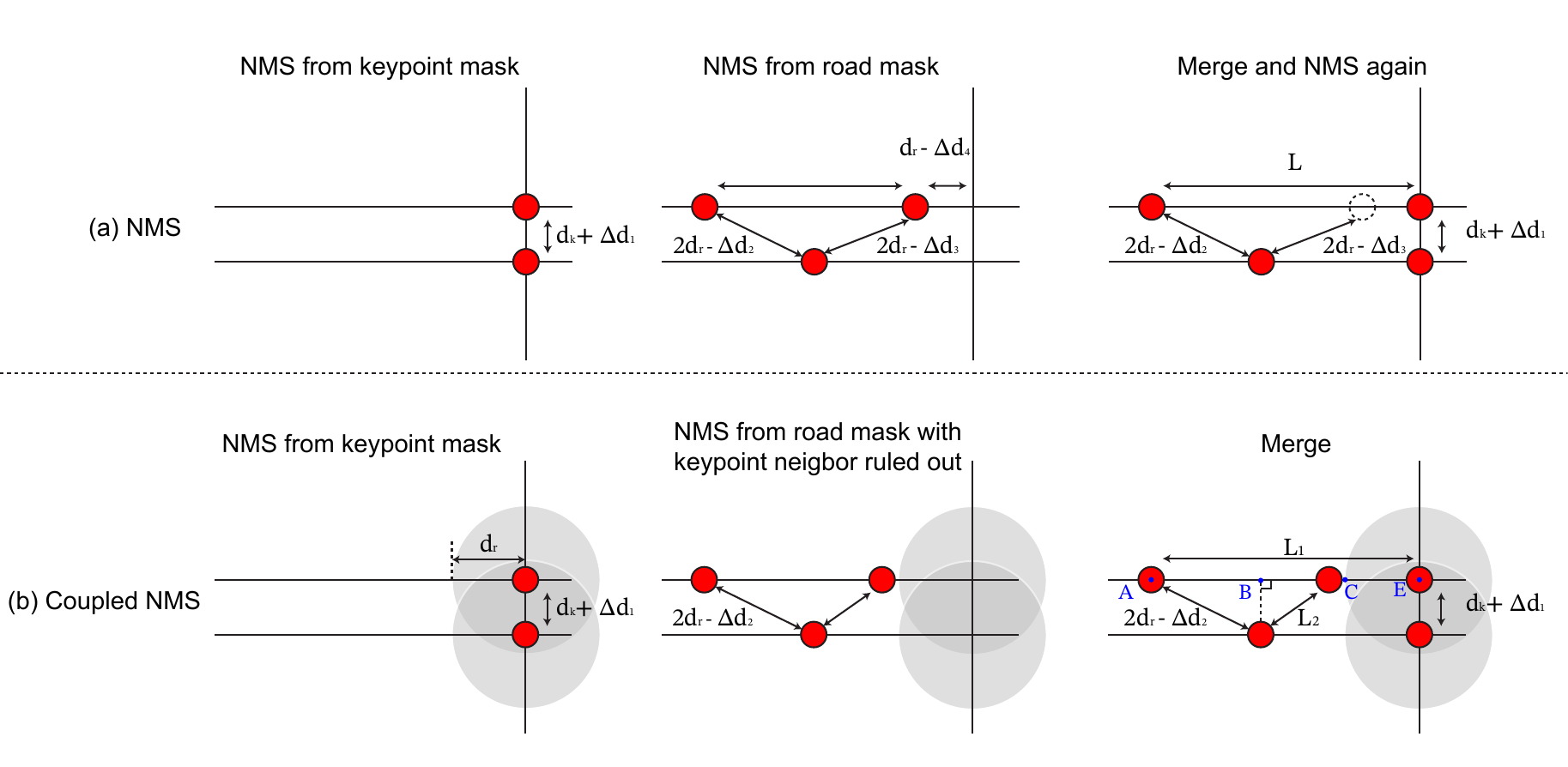}
  \caption{NMS vs. Coupled NMS. (a) Situation that will happen in NMS strategies used in \cite{hetang2024segment,yin2025towards}. (b) How coupled NMS avoids this situation.}
  \label{fig:nms_counter}
\end{figure*}

The main advantage of the coupled process is to effectively avoid the challenging failure cases encountered in the NMS strategies used in \cite{hetang2024segment,yin2025towards} as illustrated in \Cref{fig:nms_counter}.

In the general setting, we have $d_r = 2 \cdot d_k$, and $d_{nei} = 4 \cdot d_r$. The described situation is likely to occur when  $\Delta d_1, \Delta d_2, \Delta d_3, \Delta d_4 > 0$.

\begin{equation}
\begin{split}
    L = & \sqrt{(2\cdot d_r - \Delta d_2)^2 - (d_k + \Delta d_1)^2} \\
    & + \sqrt{(2\cdot d_r - \Delta d_3)^2 - (d_k + \Delta d_1)^2} \\
    & + (d_r + \Delta d_4)
\label{equ:Leq}
\end{split}
\end{equation}

Assume $\Delta d_1 = \Delta d_2 = \Delta d_3 = \Delta d_4 = \frac{d_r}{M}$ and we choose $M = \frac{20}{3}$, then we have

\begin{equation}
\begin{split}
    L  = & \sqrt{(2\cdot d_r -  \frac{d_r}{M})^2 - (\frac{d_r}{2} 
    +  \frac{d_r}{M})^2} \\
    & + \sqrt{(2\cdot d_r -  \frac{d_r}{M})^2 - (\frac{d_r}{2} +  \frac{d_r}{M})^2} + (d_r +  \frac{d_r}{M}) \nonumber \\
     = & 2 \cdot d_r \cdot \sqrt{(2-  \frac{1}{M})^2 - (\frac{1}{2} +  \frac{1}{M})^2}  + (d_r +  \frac{d_r}{M}) \nonumber \\
     = & (2 \cdot \sqrt{\frac{15}{4} - \frac{5}{M}}  + 1 +  \frac{1}{M}) \cdot d_r > 4 \cdot d_r
\label{equ:Lneq}
\end{split}
\end{equation}
This demonstrates that there exist situations where two vertices should be connected by an edge in the road network graph, yet the distance between them exceeds $4\cdot d_r$.

\begin{lemma}
    The proposed Coupled NMS effectively avoids this situation by ensuring that at least one of the conditions $L_1 < 4 \cdot d_r$ or $L_2 > d_r$ is satisfied.
\end{lemma}

\emph{Proof.} We first assume the first condition is not satisfied, i.e. $L_1 \geq 4 \cdot d_r $, then
\begin{equation}
\begin{split}
L_1 \geq 4 \cdot d_r \iff& \overline{BC} = L_1 - \overline{AB} - \overline{CE} > d_r \\ 
&(\overline{CE} = d_r \text{ and } \overline{BC} < 2 \cdot d_r - \Delta d_2) \nonumber \\
\iff& L_2 > \overline{BC} > d_r
\end{split}
\end{equation}
which means that if $L_1 > 4 \cdot d_r $, then NMS will keep one vertex between $B$ and $C$ on the road network graph.

We then assume the second condition is not satisfied, i.e. $L_2 \leq d_r $, then
\begin{equation}
\begin{split}
L_2 \leq  d_r \iff& \overline{BC} <  L_2 \leq  d_r \nonumber \\
\iff& L_1 = \overline{AB} + \overline{BC} + \overline{CE} \\
&< 2 \cdot d_r - \Delta d_2 + d_r + d_r \\
&< 4 \cdot d_r
\end{split}
\end{equation}
which means that if we couldn't find any vertex between $B$ and $C$ satisfying the NMS threshold condition, then $L_1<4 \cdot d_r$, won't exceed the pair neiborhood threshold.

\section{Candidate Edge Features from SAM Embedding}
\begin{figure}[ht]
  \centering
  \includegraphics[width=0.4\textwidth]{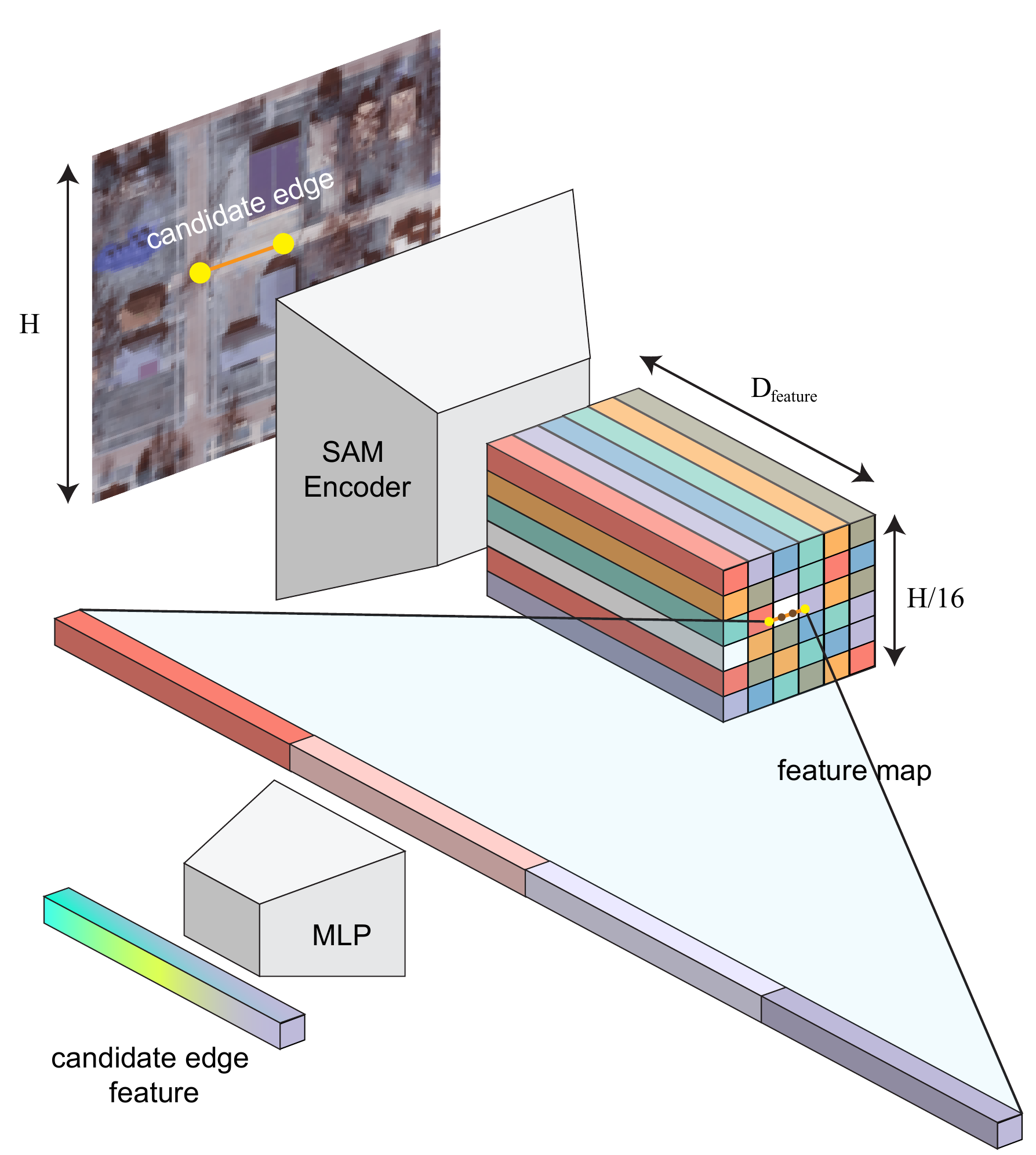}
  \caption{Process of generating candidate edge features. Intermediate points are interpolated between edge endpoints, and features are sampled at these points using a bilinear sampler. The sampled features are then concatenated and passed through an MLP to produce the final edge feature.}
  \label{fig:edgefeat}
\end{figure}
As shown in \Cref{fig:edgefeat}, we obtain the feature representation for each candidate edge by sampling visual features from the encoder feature map $F$ using bilinear interpolation at both endpoints, as well as at a series of uniformly interpolated points along the edge. This sampling strategy captures contextual cues along the potential road segment. Specifically, we extract $N_{\text{sampled}}$ feature vectors from positions evenly spaced between the two endpoints, concatenate them in order from the first to the second endpoint, and use the resulting sequence to represent the edge. The concatenated feature vector is then passed through a multilayer perceptron (MLP) to generate the final candidate edge embedding, which serves as input to the graph reasoning module for connectedness prediction.

\section{Effect of Overpass/Underpass Segmentation Head}
\label{sec:overpass_effect}
\begin{table}[ht]
  \caption{Results for models with overpass head on the City-scale dataset.}
  \small
  \label{tab:overpass_head_city}
  \centering
  \begin{adjustbox}{max width=1\textwidth}
  \begin{tabular}{l cccc}
\toprule
 & \multicolumn{4}{c}{\textbf{City-scale Dataset}} \\ 
 \cmidrule(r){2-5}
{Methods} & {Prec.↑} & {Rec.↑} & {F1↑} & {APLS↑} \\ \hline

SAM-Road~\cite{hetang2024segment}               & 90.47 & \textbf{67.69} & 77.23 & \textbf{68.37} \\ 

SAM-Road~\cite{hetang2024segment} w/ Overpass   & \textbf{92.67} & 67.52 & \textbf{77.89} & 66.82 \\ 
SAM-Road++~\cite{yin2025towards}                & 88.39 & \textbf{73.39} & \textbf{80.01} & \textbf{68.34} \\ 

SAM-Road++~\cite{yin2025towards} w/ Overpass    & \textbf{88.47} & 72.54 & 79.48 & 67.34 \\ 
\textbf{Ours}                                   & 91.09 & 75.44 & 82.37 & 68.88 \\ 
\textbf{Ours w/ Overpass}                       & \textbf{92.75}   &   \textbf{76.64}&\textbf{83.77}     &  \textbf{70.40}   \\

\bottomrule
\end{tabular}
\end{adjustbox}
\end{table}

To evaluate the effectiveness of the proposed overpass/underpass segmentation head, we compare overpass version against the non-overpass version from prior work~\cite{hetang2024segment} across three models — SAM-Road, SAM-Road++, and LineGraph2Road — on the City-scale dataset (\Cref{tab:overpass_head_city}).

The overpass/underpass head consistently enhances our framework, yielding notable gains in precision recall, TOPO-F1, and APLS. In contrast, adding it to SAM-Road and SAM-Road++ offers limited or even negative effects. This contrast highlights that the head alone is insufficient without a model capable of reasoning over structured, global context, and might introduce variability. When integrated into LineGraph2Road, however, it complements the line-graph-based connectivity reasoning by explicitly modeling multi-level intersections, enabling the network to recover complex crossing structures that other models often miss.

\section{Graph Transformer Details}
\label{sec:graph_transformer}

Let $\mathbf{X}^{(l)} = \{\mathbf{x}_1^{(l)}, \mathbf{x}_2^{(l)},...,\mathbf{x}_n^{(l)}\}$ denote the node features of the $l$-th layer and $\mathbf{X}^{(0)}$ be the candidate edge feature generated from the image feature map. For each edge from node $j$ to node $i$ in $L(G)$, we compute the multi-head attention as :

\begin{equation}
\alpha_{c,ij}^{(l)} = \mathrm{softmax} \left( 
\frac{(\mathbf{W}_{c,q}^{(l)} \mathbf{x}_i^{(l)})^\top (\mathbf{W}_{c,k}^{(l)} \mathbf{x}_j^{(l)} )}{\sqrt{d}} 
\right) \, ,
\end{equation}
where $c$ indexes the attention head, $d$ is the hidden dimension per head, and $\mathbf{W}_{c,q}^{(l)}$ and $\mathbf{W}_{c,k}^{(l)}$ are trainable parameters. After getting the multi-head attention, we compute the message aggregation for each node $i$ as:

\begin{equation}
\mathbf{h}_i^{(l)}= \big\|_{c=1}^{C} \sum_{j \in \mathcal{N}(i)} \alpha_{c,ij}^{(l)} \left( \mathbf{W}_{c,v}^{(l)} \mathbf{x}_j^{(l)}  \right) \, ,
\end{equation}
where the $\big\|$ denotes concatenation over all $C$ attention heads, and $\mathbf{W}_{c,v}^{(l)}$ is trainable parameters. The aggregated features $\mathbf{H}^{(l)} = \{\mathbf{h}_1^{(l)}, \mathbf{h}_2^{(l)},...,\mathbf{h}_n^{(l)}\}$ are passed through a residual connection and feedforward layer with layer normalization:

\begin{equation}
\begin{aligned}
\mathbf{H}^{(l+1)} &= \mathrm{LN}\!\left( \mathbf{X}^{(l)} + \mathrm{DO}\!\left(\mathbf{H}^{(l)}\right) \right), \\
\mathbf{X}^{(l+1)} &= \mathrm{LN}\!\left( \mathbf{H}^{(l+1)} + \mathrm{DO}\!\left(W_2^{(l)}\,\mathrm{ReLU}\!\left(W_1^{(l)} \mathbf{H}^{(l+1)}\right)\right) \right)
\end{aligned}
\end{equation}
where $\mathrm{LN}$ denotes Layer Normalization and $\mathrm{DO}$ denotes Dropout. $W_1^{(l)}$ and $W_2^{(l)}$ are trainable parameters. Finally, we pass the output node features from the last layer, $\mathbf{X}^{(3)}$, through a linear layer followed by a sigmoid activation to produce the binary classification output $\hat{B}$ for each candidate edge.

\section{Expressiveness of Model to Learn Structural Link Representation}
\label{sec:express}

We discuss the expressiveness of models for structural link representation learning. We first give the definition of the structural link representation.

\begin{definition}\textbf{ (Set-Isomorphism)}  
Given a graph $G = (V_G, A_G)$, two node subsets $S, S' \subseteq V_G$ are said to be \emph{set-isomorphic}, denoted $S \simeq_G S'$, if there is a permutation $\pi \in \Pi_n$ such that $\pi(A_G) = A_G$ and $\pi(S) = S'$. 
\label{def:set_isom}
\end{definition}

\begin{definition}\textbf{ (Structural Representation)}  
For $G = (V_G, A_G)$, $S \subseteq V_G$ and let $f(S, G)$ be a function that aims at learning a representation for the node set $S$.  
The function $f(S, G)$ is said to be a \emph{most expressive structural representation}, if for all $S, S' \subseteq V_G$,  
$f(S, G) = f(S', G) \iff S \simeq_G S'$. When $|S| = 2$, we refer to it as the \emph{most expressive structural link representation}, and when $|S| = 1$, we refer to it as the \emph{most expressive structural node representation}.
\label{def:structrep}

\end{definition}

 We do not consider labeling trick methods~\cite{zhang2021labeling}, as they typically prevent the model from predicting all links in a single GNN inference step, leading to reduced efficiency.


\subsection{Graph Autoencoders Fails}

For the discussion, we first give the definition of Node-most-expressive GNN~\cite{zhang2021labeling}. 

\begin{definition} \textbf{(Node-most-expressive GNN)}
A GNN is \emph{node-most-expressive} if for any graph $G = (V_G, A_G)$, $\forall i, j \in V_G$, $\text{GNN}(i, A_G) = \text{GNN}(j, A_G) \Leftrightarrow {i} \simeq_G {j}$.
\label{def:mostgnn}
\end{definition}

As proved in~\cite{zhang2021labeling}, GAEs cannot learn a structural link representation, even when equipped with a \emph{node-most-expressive GNN} that learns a structural node representation. This is demonstrated by presenting examples of multi-node representation learning problems involving more than two nodes, showing that directly aggregating node representations from a GNN does not yield a structural representation for node sets.

\begin{figure}[ht]
  \centering
  \includegraphics[width=0.3\textwidth]{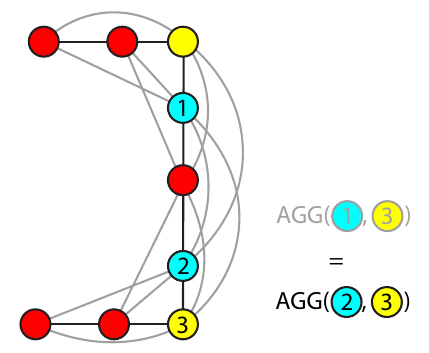}
  \caption{directly aggregating node representations from a GNN does not yield a structural representation for node sets}
  \label{fig:GAE}
\end{figure}

Here, we provide examples based on the road network graph to illustrate that this issue also arises in the original graph we construct for connectedness prediction with node-most-expressive GNN as shown in \Cref{fig:GAE}.

\begin{assumption} If $\{i\} \simeq_G \{j\}$, $i,j\in V_G$, the initial node feature embeddings for them input to the GAE are necessarily identical.
\end{assumption}

\begin{figure}[ht]
  \centering
  \includegraphics[width=1\textwidth]{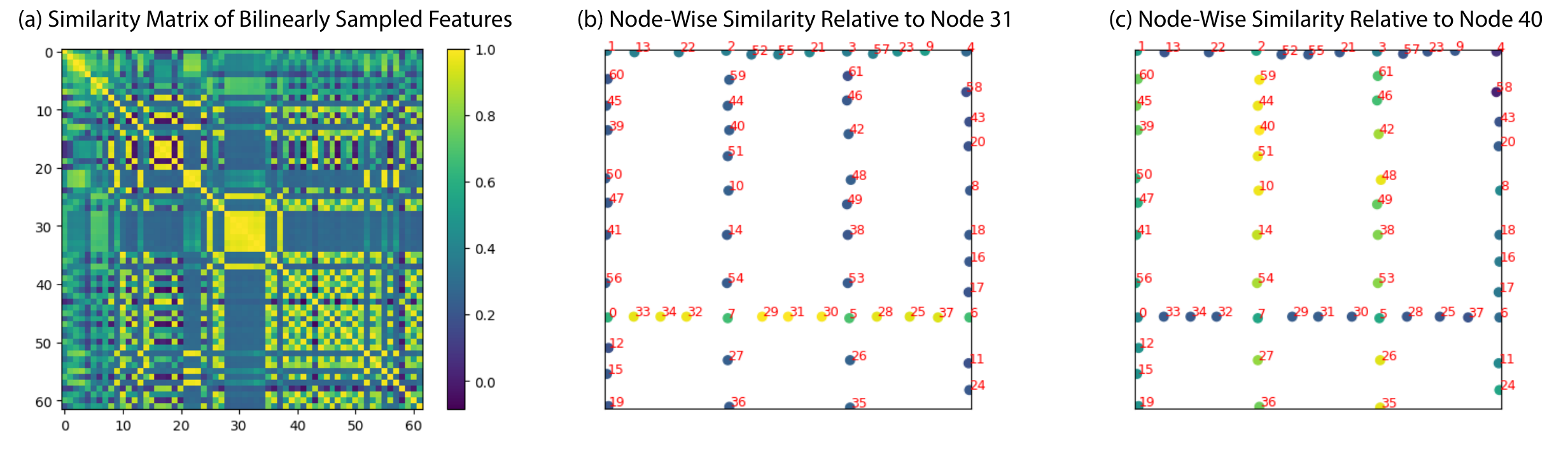}
  \caption{Pairwise similarity of bilinearly sampled features across nodes in the graph. (a) The overall similarity matrix shows the pairwise feature similarities for all nodes. (b) Node-wise similarity relative to node 31. (c) Node-wise similarity relative to node 40.}
  \label{fig:sim}
\end{figure}

While this assumption does not strictly hold in practice, it is a reasonable approximation supported by empirical evidence. Specifically, bilinearly sampled features corresponding to isomorphic nodes consistently exhibit high cosine similarity, as demonstrated in \Cref{fig:sim}.

In this example, we have $\{1\} \simeq_G \{2\}$, then by \Cref{def:mostgnn}, we will have $\text{GNN}(\{1\}, A_G) = \text{GNN}(\{2\}, A_G)$, and further we have  
\begin{equation}
    AGG(\text{GNN}(\{1\}, A_G), \text{GNN}(\{3\}, A_G)) = AGG(\text{GNN}(\{2\}, A_G), \text{GNN}(\{3\}, A_G))
\end{equation}

However, $\{1,3\}$ and $\{2,3\}$ is not \emph{set-isomorphic}, so the aggregated most expressive structural node representation is not a  most expressive structural link representation.

\subsection{Line Graph for Structural Link Representation}

Now, we discuss the ability of Node-most-expressive GNN to learn structural link representation by line graph conversion.

\begin{theorem}\textbf{(Whitney's Theorem)} If two connected graphs have isomorphic line graphs, then the original graphs are isomorphic---\emph{except for the pair} $K_3$ and $K_{1,3}$ \cite{whitney1932congruent}
.\label{sec:whitney}
\end{theorem}

\begin{lemma}
    Given a connected graph $G = (V_G, A_G)$ and its corresponding line graph $L(G) = (V_{L(G)}, A_{L(G)})$, we have $i,j \in V_G, A_{ij}=1$, then $ij \in V_{L(G)}$. For link $(i,j)$ in $G$, $f((i,j), G) = \text{GNN}(ij, A_{L(G)}) $ is the most expressive structural link representation, if the GNN is \emph{node-most-expressive} and $G \notin \{K_3 , K_{1,3}\}$.
\end{lemma}

\emph{Proof.} According to \Cref{def:structrep}, as long as for all $\{i,j\},  \{k,l\} \subseteq V_G$ and $A_{ij} = A_{kl} = 1$,  
$ \text{GNN}(ij, A_{L(G)}) =  \text{GNN}(kl, A_{L(G)}) \iff \{i,j\} \simeq_G \{k,l\}$ , then $f((i,j), G) = \text{GNN}(ij, A_{L(G)}) $ is the most expressive structural link representation.

\begin{equation}
\begin{split}
    &\text{GNN}(ij, A_{L(G)}) = \text{GNN}(kl, A_{L(G)}), \text{ }  ij, kl \in V_{L(G)}\nonumber \\
    &\iff ij \simeq_{L(G)} kl \quad \\
    & \quad \quad \quad \quad \text{(by \Cref{def:mostgnn} and GNN is \emph{node-most-expressive})} \nonumber \\
    &\iff \exists\, \pi \in \Pi_m \text{ s.t. } \pi(A_{L(G)}) = A_{L(G)} \text{ and } \pi(ij) = kl \\
    & \quad \quad \quad \quad \text{(by \Cref{def:set_isom})} \nonumber \\
    &\iff \exists\, \pi' \in \Pi_n \text{ s.t. } \pi'(A_G) = A_G \text{ and } \pi'(\{i,j\}) = \{k,l\} \nonumber \\
     & \quad \quad \quad \quad \text{(by \Cref{sec:whitney} and  graph $G \notin \{K_3 , K_{1,3}\}$)} \nonumber \\
    &\iff\{i,j\} \simeq_G \{k,l\}
\end{split}
\end{equation}
This concludes the proof.

While triangles ($K_3$) and 3-star graphs ($K_{1,3}$) are theoretical exceptions to line-graph isomorphism, they almost never occur as isolated components in real road networks. Genuine instances—such as a standalone roundabout with no connecting roads or an isolated three-way split of very short segments—are exceedingly rare and largely artificial. To confirm this empirically, we scanned two large datasets with \Cref{alg:count_isolated}.
\begin{algorithm}[ht]
\caption{Counting Isolated $K_3$ and $K_{1,3}$ Components}
\label{alg:count_isolated}
\begin{algorithmic}[1]
\STATE $count\_K_3 \gets 0$, $count\_K_{13} \gets 0$
\FOR{each $id$ in \texttt{ids}}
    \STATE $G \gets$ \texttt{load\_graph}($id$)
    \FOR{each $comp$ in \texttt{connected\_components}($G$)}
        \IF{$|comp| = 3$ \textbf{and} every node has degree $2$ within $comp$}
            \STATE $count\_K_3 \gets count\_K_3 + 1$
        \ELSIF{$|comp| = 4$ \textbf{and} one node has degree $3$ and three nodes have degree $1$ within $comp$}
            \STATE $count\_K_{13} \gets count\_K_{13} + 1$
        \ENDIF
    \ENDFOR
\ENDFOR
\end{algorithmic}
\end{algorithm}

This yields: {City-Scale} (over 755\,km$^2$) contains 1 instance of $K_3$ and 10 instances of $K_{1,3}$, while {SpaceNet} (over 407\,km$^2$) contains 0 instances of $K_3$ and 9 of $K_{1,3}$. These counts confirm that truly isolated $K_3$ and $K_{1,3}$ patterns are negligible in practice, indicating that the theoretical exception has minimal impact on our method’s validity. 

\subsection{Comparison with Standard GNN Baselines as Graph Autoencoders on the Line Graph}
\label{sec:gnn}


\begin{table}[ht]
  \caption{Results for different GNN models with overpass head on the City-scale dataset.}
  \small
  \label{tab:overpass_city_gnn}
  \centering
  \begin{adjustbox}{max width=1\textwidth}
  \begin{tabular}{l ccccc}
\toprule
 & \multicolumn{4}{c}{\textbf{City-scale Dataset}} \\ 
 \cmidrule(r){2-5}
{GNN (w/ Overpass )} & {Prec.↑} & {Rec.↑} & {F1↑} & {APLS↑} & {Time per Image (s)} \\ \hline

GCN~\cite{kipf2016semi}           & {92.92} & 75.67 & 83.24 & {70.14} & 10.51\\ 


GAT~\cite{velivckovic2017graph}  & {92.97} & {75.85} & {83.34} & 70.22 & 10.74\\ 


SAGE~\cite{hamilton2017inductive}                & 93.59 & {72.44} & {81.37} & {64.13} & 10.64\\ 


GIN~\cite{xu2018powerful}   & {92.81} & 76.13 & 83.49 & \textbf{70.54} & {10.48} \\ 

\textbf{Ours (Graph Transformer\cite{shi2020masked})}                       & {92.75}   &   {76.64}&\textbf{83.77}     &  70.40 & 10.51 \\

\bottomrule

 Inference time for $5 \times 5$ sliding windows
\end{tabular}
\end{adjustbox}
\end{table}

Table~\ref{tab:overpass_city_gnn} reports the performance of different GNN-based graph autoencoders (GCN, GAT, GraphSAGE, and GIN) using the same line-graph representation and training protocol with an overpass prediction head on the City-scale dataset. The comparison shows that while the line-graph representation itself provides a strong foundation, our Graph Transformer consistently achieves the best overall F1 score and recall, while maintaining comparable computational cost to the simpler GNN variants. We adopt the Graph Transformer architecture~\cite{shi2020masked} to preserve a transformer-based design consistent with~\cite{hetang2024segment,yin2025towards}, ensuring that the observed performance gains stem from enhanced graph reasoning rather than differences in model class.

\section{Pipeline for Preprocessing Training Data}

\label{sec:preprocess}

\begin{algorithm}
\caption{Iterative Graph Refinement for Road Network Overpass/Underpass Detection and Adjustment}
\label{alg:road_refine}
\begin{algorithmic}[1]
\STATE \texttt{// $\tau$: intersection threshold,\ $\gamma$: merge gap,\ $\alpha$: step scale}
\STATE \texttt{// $T$: max iterations,\ $P$: merge period,\ $\varepsilon$: tolerance}

\STATE $G \leftarrow G_0$
\FOR{$t = 1$ \TO $T$}
    \STATE Build spatial index over edges; $\Delta \leftarrow \emptyset$
    \FOR{each intersecting edge pair $(a,b),(c,d)$ without shared endpoints}
        \FOR{each endpoint $v \in \{a,b,c,d\}$}
            \STATE $d \leftarrow \|v - p\|$
            \IF{$d < \tau$}
                \STATE $\hat{u} \leftarrow$ unit vector from $v$ to its neighbor
                \STATE $\Delta[v] \mathrel{+}= ((\tau - d)/\tau) \cdot \hat{u}$
            \ENDIF
        \ENDFOR
    \ENDFOR

    \IF{$t \bmod P = 0$}
        \STATE \texttt{// Contract stacked nodes}
        \FOR{each node $k$ with $\deg(k)=2$ and gap$(k) < \gamma$}
            \STATE Connect its two neighbors and remove $k$
        \ENDFOR
    \ELSE
        \STATE $moved \leftarrow 0$
        \FOR{each node $v$ with adjustment $\mathbf{a} = \Delta[v]$}
            \IF{$\|\mathbf{a}\| > \varepsilon$}
                \STATE $\hat{a} \leftarrow \mathbf{a} / \|\mathbf{a}\|$
                \STATE $v' \leftarrow v + \alpha \hat{a}$
                \IF{$v'$ not overlapping other nodes}
                    \STATE Move $v$ to $v'$, $moved \leftarrow moved + 1$
                \ENDIF
            \ENDIF
        \ENDFOR
        \IF{$moved = 0$}
            \STATE \textbf{break} \texttt{// converged}
        \ENDIF
    \ENDIF
\ENDFOR

\RETURN $G$
\end{algorithmic}
\end{algorithm}

\Cref{alg:road_refine} outlines the iterative graph refinement process used to disentangle stacked road intersections and identify potential overpass/underpass crossing structures. Starting from an initial road graph $G_0$, the algorithm repeatedly adjusts node positions based on local geometric conflicts detected through edge intersections. For each intersecting pair of edges, nodes within a proximity threshold $\tau$ are shifted along their incident directions to reduce overlap. Every $P$ iterations, degree-2 nodes forming near-parallel, closely spaced connections (gap $<\gamma$) are merged to collapse redundant stacked segments. The iterative relocation continues until all node displacements fall below the tolerance $\varepsilon$, yielding a refined graph $G$ in which true overpasses/underpasses crossing remain geometrically separated while false planar intersections are resolved.

\begin{figure}[ht]
  \centering
  \includegraphics[width=0.75\textwidth]{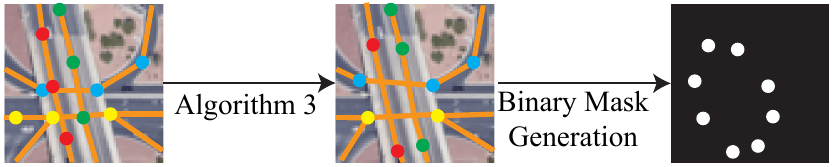}
  \caption{Overpass/underpass masks generation process}
  \label{fig:overpass_mask}
\end{figure}

To finally generate the overpass/underpass masks, we focus on the endpoints of edges identified as geometrically separated yet intersecting overpass/underpass crossings by \Cref{alg:road_refine}, and convert them into binary masks that serve as supervision targets for our model.

\begin{figure*}[ht]
  \centering
  \includegraphics[width=1\textwidth]{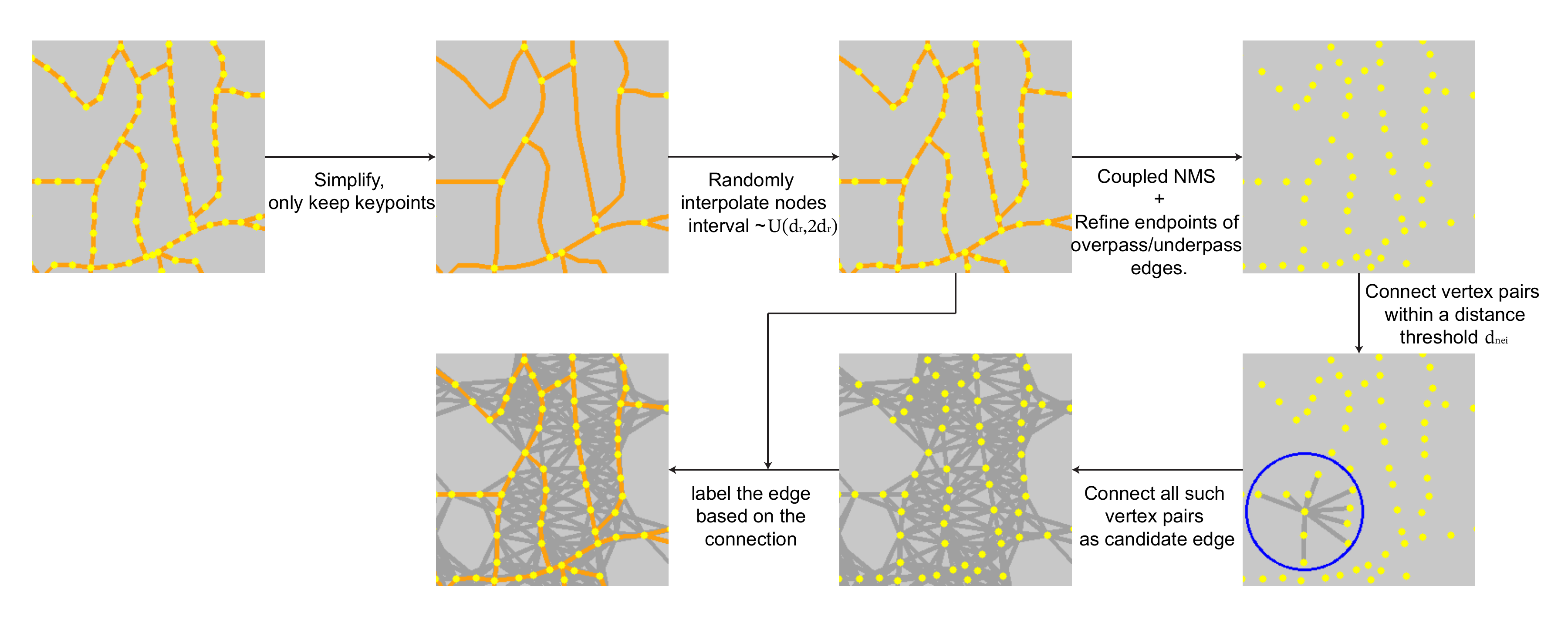}
  \caption{Pipeline for preprocessing the original road network graph into the final graph used for classification.}
  \label{fig:preprocess}
\end{figure*}

\Cref{fig:preprocess} illustrates a process for generating candidate edges with label for training data. Initially, only the keypoints are retained, and additional nodes are randomly interpolated between them with intervals sampled from a uniform distribution $U(d_r, 2d_r)$ according to \Cref{alg:sampling_distances}. Coupled NMS is then applied to both the keypoints and interpolated nodes to refine the set of vertices. \Cref{alg:road_refine} is used to detect the endpoints of edges involved in overpass/underpass crossings and adjust them to suitable locations.Vertex pairs within a specified distance threshold \( d_{\text{nei}} \) are connected, forming potential edges. These pairs are considered as candidate edges, which are subsequently labeled based on the connection criteria.

\begin{algorithm}
\caption{Randomly Interpolate Nodes Along a Line}
\label{alg:sampling_distances}
\begin{algorithmic}[1]
\STATE \texttt{// $d_r$: road suppression radius}
\STATE $D \leftarrow \emptyset$ \texttt{//List of sampled distances}
\STATE $p \leftarrow 0$ \texttt{//Current position}
\STATE $L \leftarrow$ length of the line

\WHILE{$p + 2 \cdot d_r - 1 < L$}
    \STATE Sample $u \sim \mathcal{U}(d_r,\ 2 \cdot d_r - 1)$
    \IF{$p + u + d_r - 1 < L$}
        \STATE $p \leftarrow p + u$
        \STATE Append $p$ to $D$
    \ENDIF
\ENDWHILE
\RETURN $D$
\end{algorithmic}
\end{algorithm}

\section{Additional Visual Comparison}
\label{sec:addvis}

\subsection{More Qualitative Comparison on City-scale Dataset}
\label{sec:addviscity}

\begin{figure*}
    \centering
    \includegraphics[width=0.95\linewidth]{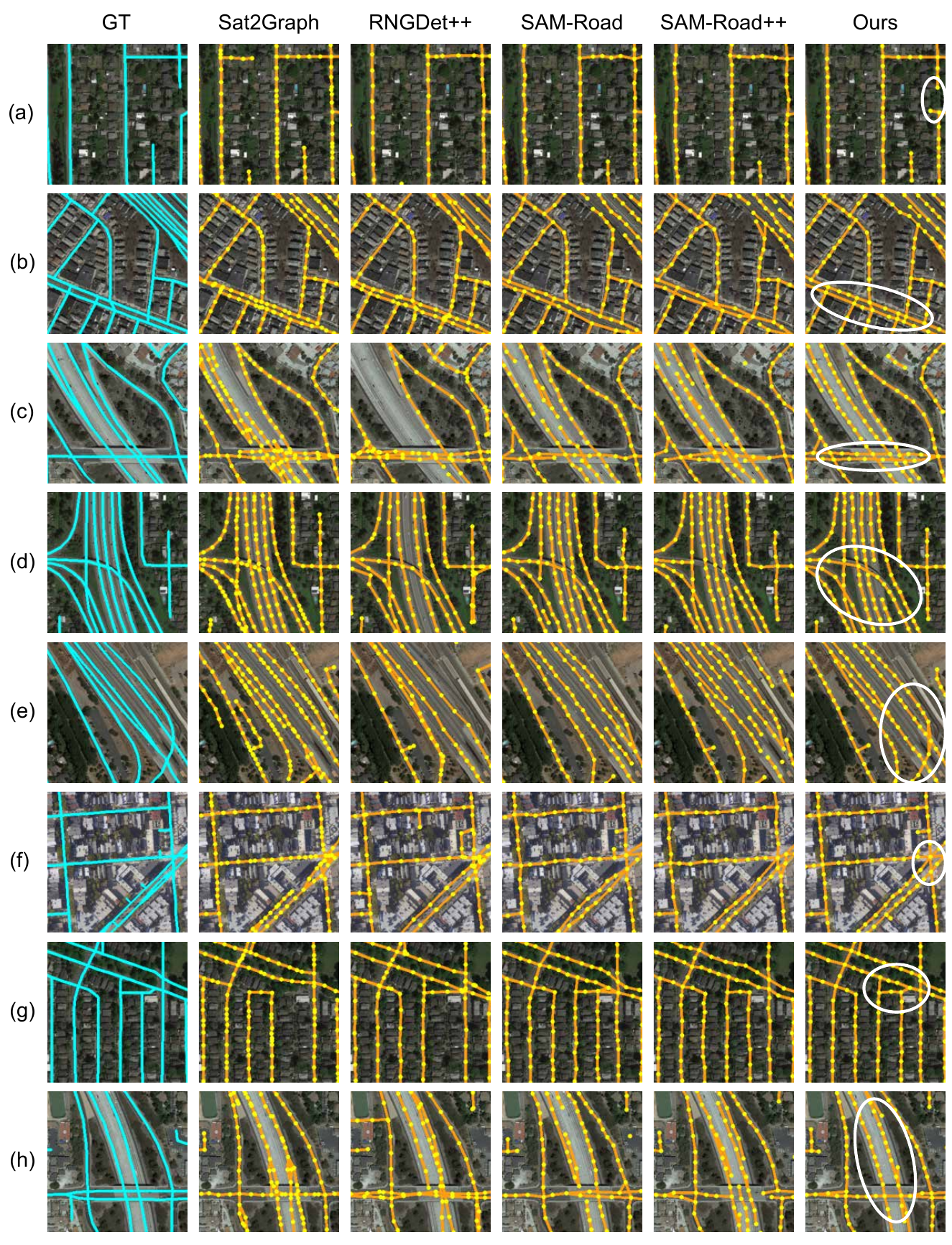}
    \caption{Additional visualizations showing the model’s ability to capture complex topologies, including overpasses(panels c–e). While some imperfections remain, the model often avoids false intersections and preserves overall road geometry better than existing methods.}
    \label{fig:additional}
\end{figure*}

We provide additional visualizations in \Cref{fig:additional} to further demonstrate our model’s ability to capture complex topological structures. \Cref{fig:additional}c,d,e show our model's promising performance for overpass scenarios. In \Cref{fig:additional}c, two parallel lateral highways pass over three vertical roads, and the model correctly avoids predicting false intersections between the lower lateral highway and the vertical roads. A similar result is observed in \Cref{fig:additional}d, where the curved highway is appropriately separated from the vertical highway. Nonetheless, some limitations remain: in \Cref{fig:additional}c, the upper lateral highway is fragmented; in \Cref{fig:additional}d, the rightmost vertical lines are partially broken. In \Cref{fig:additional}e, the curved highway is incorrectly connected to nearby lanes. Despite these imperfections, our model captures the overall geometry more accurately than existing methods, which often fail to preserve the curved structure altogether.

\begin{figure*}
  \centering
  \includegraphics[width=0.95\textwidth]{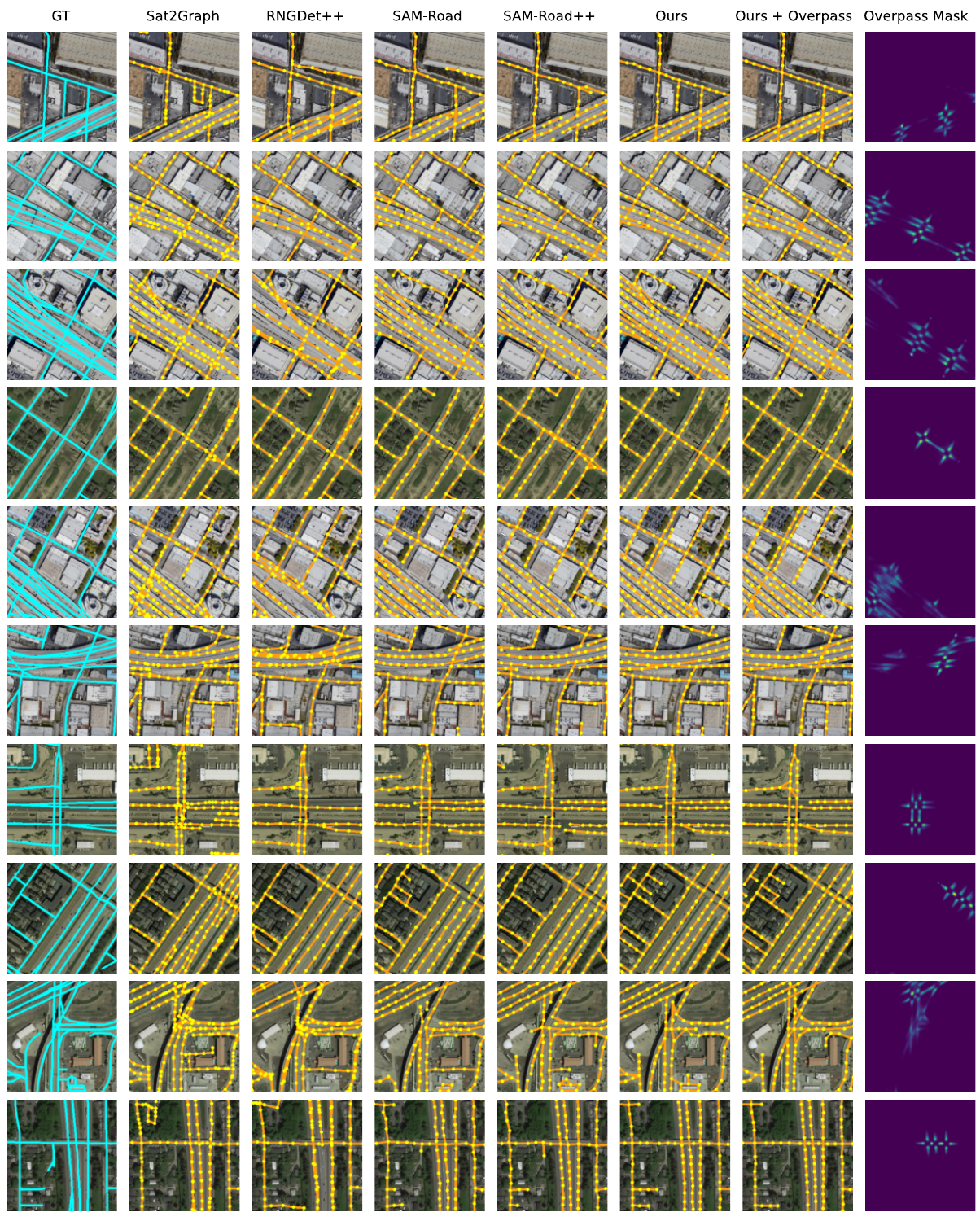}
    \caption{Additional visualized road network graph predictions of our method and other baseline methods as well as the predicted overpass/underpass mask on City-Scale dataset, illustrating performance on overpass and underpass intersections.}

  \label{fig:vizoverpassapp}
\end{figure*}

Additional visualizations in \Cref{fig:vizoverpassapp} further highlight our overpass variant’s capability to accurately capture complex overpass and underpass crossings.

\subsection{Failure Cases of Global-Scale Dataset}
\label{sec:failureglobal}

\begin{figure*}
    \centering
    \includegraphics[width=1\linewidth]{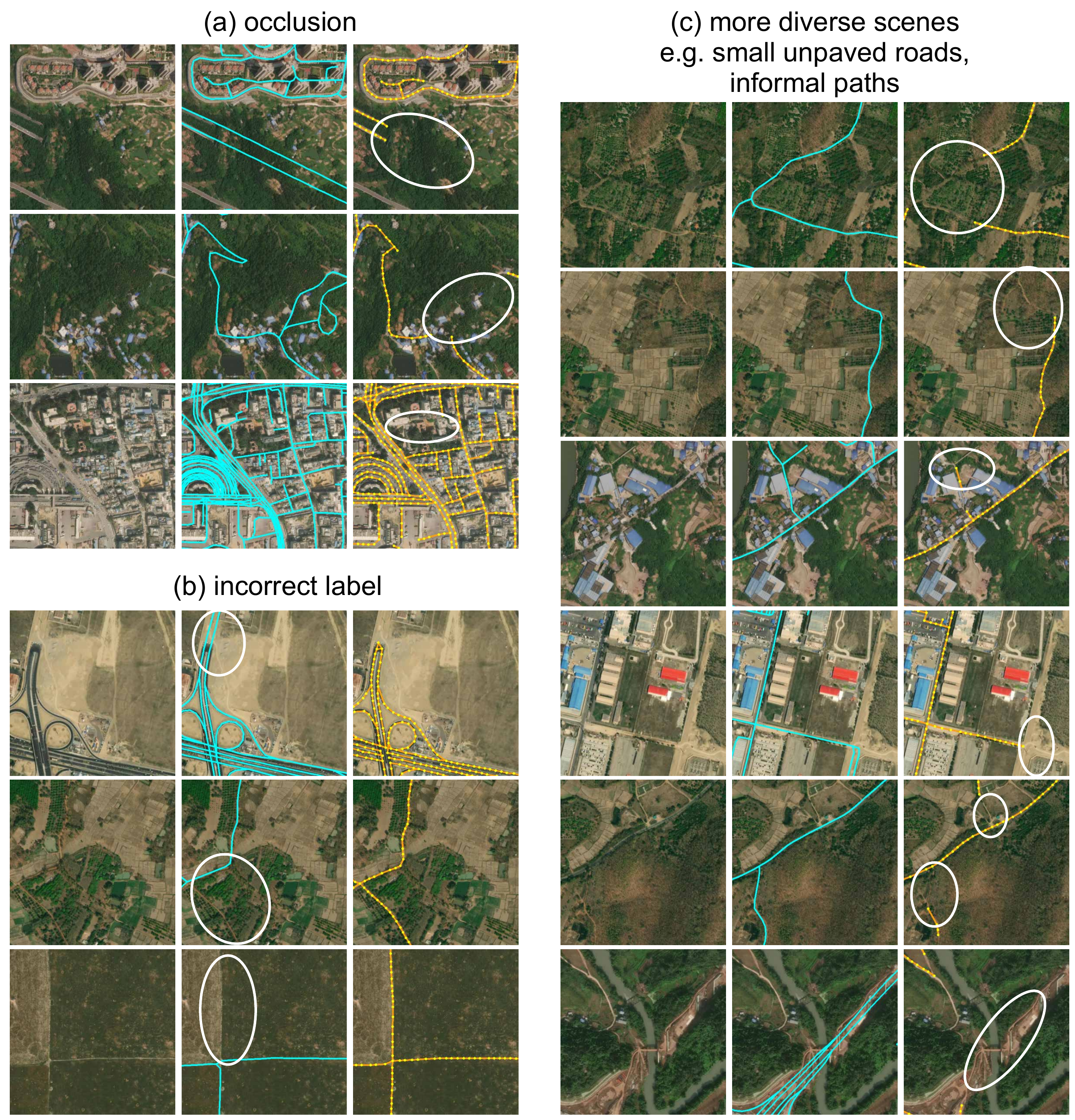}
    \caption{Representative failure cases on the Global-scale dataset. (a) Missed roads due to occlusions from tunnels, vegetation, or tall buildings. (b) Incorrect or missing labels in the test set. (c) Missed detections of small unpaved or informal roads, reflecting domain shifts across diverse geographic regions.}
    \label{fig:failurecase}
\end{figure*}

We present representative failure cases from the Global-scale dataset in \Cref{fig:failurecase}, which we categorize into three types. The first involves occlusions caused by tunnels, dense vegetation, or tall buildings, as shown in \Cref{fig:failurecase}a. The second type stems from incorrect or missing labels in the test set as shown in \Cref{fig:failurecase}b, which is an inevitable issue in large-scale datasets and can lead to an underestimation of model performance. The third category includes small unpaved roads or informal paths as shown in \Cref{fig:failurecase}c. This case highlights a broader challenge of domain shift and annotation inconsistency in large-scale, globally sourced datasets. Specifically, the failure involves small unpaved or informal roads, which are difficult to detect for several reasons. First, the dataset combines imagery from diverse geographic and socioeconomic regions. In rural or remote areas, unpaved roads and informal paths are often annotated as valid roads. However, in urban areas or even other rural areas, similar features are either unlabeled or inconsistently labeled. For example, the upper part of 5th row of \Cref{fig:failurecase}c has two unpaved small roads without labels according to the ground truth. This lack of consistent ground truth introduces ambiguity during training, where visually similar features may be considered "road" in one area and "non-road" in another. Second, small and unpaved roads often have limited contrast with the surrounding environment, especially in agricultural or arid regions, making it difficult for the model to detect them without clear boundaries. The model struggles to distinguish these paths from the background without sufficient training data. Third, our current model is trained on the full dataset without explicit domain adaptation techniques. As a result, it lacks the ability to adapt to regional visual and labeling differences.

\subsection{Connectivity Recovery under Occlusions}

\begin{figure}[ht]
  \centering
  \includegraphics[width=1\textwidth]{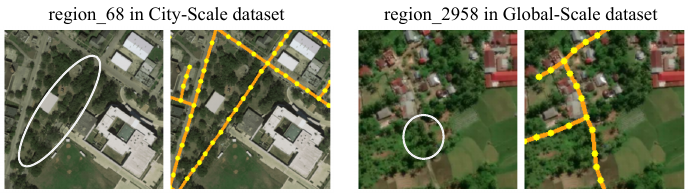}
  \caption{Examples where our method successfully recovers connectivity for tree-occluded roads.}
  \label{fig:occlusion}
\end{figure}

Graph reasoning enables the recovery of connectivity across occluded segments by propagating messages from nearby unoccluded nodes. We illustrate this capability with examples of tree-occluded roads in Fig.~\ref{fig:occlusion} for both City-Scale and Global-Scale dataset.

\section{Sliding Window}
\label{sec:sliding}

During inference, we use a $16 \times 16$ sliding window for the City-scale and Global-scale datasets, while the SpaceNet dataset is processed without windowing. To assess the impact of window sparsity, we evaluate performance using a reduced $5 \times 5$ sliding window configuration on $2048 \times 2048$ images, where each window has a size of $512 \times 512$. The $5 \times 5$ setting is the smallest number of sliding windows that fully covers the image with an overlap greater than the neighborhood threshold $d_{\text{nei}} = 64$. As shown in \Cref{tab:sliding}, reducing the number of windows has minimal impact on overall performance. In fact, the APLS score slightly improves under the $5 \times 5$ configuration for the non-overpass variant on the City-scale dataset and the overpass variant on the Global-scale dataset. In all cases, both metrics under the $5 \times 5$ setting still exceed those of all baseline methods.

\begin{table*}[ht]
  \caption{Comparison with $5\times5$ and $16\times16$ sliding windows on City-scale and Global-scale datasets.}
  \label{tab:sliding}
  \centering
  \begin{tabular}{l cccc cccc}
\toprule
 & \multicolumn{4}{c}{\textbf{City-scale Dataset}} & \multicolumn{4}{c}{\textbf{Global-scale(In-Domain)}}           \\ \cmidrule(r){2-5} \cmidrule(r){6-9}

{Methods}           & {Prec.↑} & {Rec.↑} & {F1↑} & {APLS↑} & {Prec.↑} & {Rec.↑} & {F1↑} & {APLS↑} \\ \hline
SAM-Road  ~\cite{hetang2024segment}               & 90.47           & 67.69          & 77.23         & 68.37          & {91.93}           & 45.64          & 59.80         & 59.08          \\ 
SAM-Road++ ~\cite{yin2025towards}                & 88.39       & 73.39 & 80.01 & 68.34        & 88.95       & 49.27 & 62.33 & 62.19       \\ 
\textbf{Ours}  $5\times 5$               & 90.28       & {75.93} & 82.35 & {69.05}        & {90.07} & 50.81    & 64.03         & 66.58       \\ 
\textbf{Ours}  $16\times 16$              & {91.09}   &   {75.44}&{82.37}     &  {68.88}  &    {90.54}   &   {51.12}&{64.43}     &  {67.69}            \\ 

\textbf{Ours w/ overpass}  $5\times 5$               & 92.54       & {75.60} & 83.01 & {70.58}        &  87.31    & 53.06         & {65.26} & 69.66       \\ 
\textbf{Ours w/ overpass}  $16\times 16$              & {92.75}   &   {76.64}&{83.77}     &  {70.40}     & 89.65  & 50.85 & 64.05 & {68.70}            \\ 
\bottomrule
\end{tabular}
\end{table*}

\section{Inference Time}
\label{sec:inftime}

We evaluate the inference efficiency of our model on the SpaceNet, City-scale, and Global-scale datasets using a single NVIDIA A100 GPU. As summarized in \Cref{tab:inferencetime}, inference on small image tiles, such as those in SpaceNet, is computationally efficient. In contrast, the City-scale and Global-scale datasets involve larger input image size, requiring the use of sliding windows and thereby incurring higher computational costs. However, when employing a reduced $5 \times 5$ sliding window configuration—which, as shown in \Cref{sec:sliding}, maintains comparable performance—inference time is significantly reduced. This demonstrates the practicality of our approach for large-scale applications without compromising accuracy.

\begin{table}[ht]
  \caption{Inference times for our model on three datasets using a single NVIDIA A100 GPU}
  \label{tab:inferencetime}
  \centering
  \begin{adjustbox}{max width=1\textwidth}
  \begin{tabular}{lccccc}
    \toprule
    Dataset & Image Size & \# Test Images & Sliding Windows & Total Time (s) & Time per Image (s) \\
    \midrule
    SpaceNet & $400$ & 382 & No & 115.87 & 0.30 \\
        \hline

    \multirow{2}{*}{City-scale} & \multirow{2}{*}{$2048$} & \multirow{2}{*}{27} & $16 \times 16$ & 2354.27 & 87.19 \\
     &  &  & $5 \times 5$ & 283.97 & 10.51 \\
         \hline

    \multirow{2}{*}{Global-scale} & \multirow{2}{*}{$2048$} & \multirow{2}{*}{624} & $16 \times 16$ & 47527.00 & 67.16 \\
     &  &  & $5 \times 5$ & 5012.74 & 8.03 \\
    \bottomrule
  \end{tabular}
  \end{adjustbox}
\end{table}

We further compare the inference time of our model with SAM-Road~\cite{hetang2024segment} and SAM-Road++\cite{yin2025towards} using a single NVIDIA A100 GPU. These two methods have been shown to be significantly more efficient than earlier approaches such as Sat2Graph\cite{he2020sat2graph} and RNGDet++~\cite{xu2023rngdet++}, with SAM-Road achieving up to 80× speedup. In our comparison, we use the $5 \times 5$ sliding window configuration for our model, while reporting the results of SAM-Road and SAM-Road++ under their standard $16 \times 16$ sliding window setting. This is because their performance degrades with fewer windows; for example, the APLS score of SAM-Road on the City-scale dataset drops from 68.37 to 67.21 when reducing the window configuration from $16 \times 16$ to $8 \times 8$. With $5 \times 5$ sliding window, our method achieves faster inference on both the City-scale and SpaceNet datasets as shown in \Cref{tab:inference_comparison}, at the same time, have better performance.

\begin{table}[ht]
  \caption{Total inference time comparison on City-scale and SpaceNet dataset}
  \label{tab:inference_comparison}
  \centering
  \begin{tabular}{lcc}
    \toprule
    Method & City-scale Dataset & SpaceNet Dataset \\
    \midrule
    SAM-Road & 551.19 s & 918.47 s \\ 
    SAM-Road++ & 576.51 s &  752.25 s \\
    Our  & 283.97 s & 115.87 s \\
    \bottomrule
  \end{tabular}
\end{table}


\section{Effect of Joint Training with LineGraph2Road on SAM Segmentation Performance}

To assess whether the relation reasoning module in LineGraph2Road can guide the training of segmentation features, we compare the performance of the SAM when trained independently versus when jointly trained within the LineGraph2Road framework. We evaluate both approaches on road and keypoint segmentation tasks using the City-scale and SpaceNet dataset. We use mIOU to evaluate the performance, which is defined by the overlap between the predicted segmentation and the ground truth, divided by the total area covered by the union of the two. As shown in \Cref{tab:miou}, integrating SAM into the end-to-end training pipeline of LineGraph2Road yields a slight improvement in segmentation performance. 

\begin{table}[ht]
  \caption{Comparison of segmentation performance (mIoU) on the City-scale and SpaceNet dataset. }
  \label{tab:miou}
  \centering
  \begin{adjustbox}{max width=1\textwidth}
  \begin{tabular}{lcccc}
  
    \toprule
     & \multicolumn{2}{c}{\textbf{City-scale Dataset}} & \multicolumn{2}{c}{\textbf{SpaceNet Dataset}}           \\ 
    \cmidrule(r){2-3} \cmidrule(r){4-5}
    {Methods} & {Road } & {Keypoint} & {Road } & {Keypoint} \\
    \midrule
    SAM (independent training) & 42.17 & 26.35 & 46.07& 28.38\\
    SAM (with LineGraph2Road) & \textbf{42.54} & \textbf{27.37} & \textbf{46.27}& \textbf{28.51} \\
    \bottomrule
  \end{tabular}
  \end{adjustbox}
\end{table}

While segmentation quality remains stable, joint training provides a clear advantage in graph-level reasoning—an essential capability for downstream topological tasks. 
In contrast, a SAM encoder-decoder trained in isolation cannot produce meaningful feature for graph structure inference.

To further examine this effect, we conduct an ablation study comparing joint and sequential training strategies. 
In the sequential setup, the SAM encoder-decoder is first trained for segmentation and then frozen, while the graph transformer on the line graph is trained separately. 
As reported in \Cref{tab:joint_training}, this decoupled approach significantly reduces performance on the City-scale dataset.

\begin{table}[ht]
\centering
\caption{Effect of joint versus separate training on the City-scale dataset.}
\label{tab:joint_training}
\small
\begin{adjustbox}{max width=1\textwidth}
\begin{tabular}{lcccc}
\toprule
\textbf{Model} & \textbf{Prec.↑} & \textbf{Rec.↑} & \textbf{F1↑} & \textbf{APLS↑} \\
\midrule
Train separately & 92.18 & 56.82 & 69.82 & 46.42 \\
Train jointly    & \textbf{91.09} & \textbf{75.44} & \textbf{82.37} & \textbf{68.88} \\
\bottomrule
\end{tabular}
\end{adjustbox}
\end{table}

These results demonstrate that joint training enables the encoder to adapt its features for more effective topology extraction, achieving substantial improvements in graph accuracy without compromising segmentation mIoU.

\section{Ablation Study on Neighborhood Threshold}

\begin{table}[ht]
  \caption{Comparison between different threshold configurations.}
  \label{tab:comparison_nodes}
  \small
  \centering
  \begin{tabular}{l cccc}
\toprule

{Threshold} & {Prec.↑} & {Rec.↑} & {F1↑} & {APLS↑} \\ \hline

32              & 91.43 & {75.93} & {82.76} & 67.41 \\ 
64  & 92.75 & \textbf{76.64} & \textbf{83.77} & \textbf{70.40} \\ 
128             & \textbf{93.64} & 69.69 & 79.59 & 67.34 \\ 

\bottomrule
\end{tabular}
\end{table}
Table~\ref{tab:comparison_nodes} compares different neighborhood thresholds $d_{\text{nei}}$ for graph construction. Increasing the threshold from 32 to 64 consistently improves Recall, F1 score, and APLS, indicating better connectivity recovery and global path consistency. Although the 128 setting achieves the highest Precision (93.64), it suffers from noticeably lower Recall and F1, suggesting that overly large neighborhoods may introduce conservative connectivity that misses valid edges.

Overall, $d_{\text{nei}}=64$ achieves the best balance between Precision and Recall, yielding the highest F1 and APLS scores. Therefore, we adopt $d_{\text{nei}}=64$ as the best practical choice for graph construction.

\section{Comparison with Readouts of GNN on Original Graph}
\label{sec:app_readout}

To verify that the performance gain of the line-graph formulation is not merely due to the choice of edge readout, we compare it against several original-graph link-prediction readouts under the same sparse candidate graph and visual feature setting. In these original-graph variants, the Graph Transformer operates on the original vertex graph, and each candidate edge is classified using the embeddings of its two endpoint vertices. We evaluate several commonly used readout functions, including concatenation followed by an MLP, dot-product scoring, bilinear scoring, and a Hadamard product followed by an MLP.

As shown in \Cref{tab:readout}, stronger readout functions improve the original-graph baseline, with bilinear scoring achieving the best performance among the original-graph variants. Nevertheless, the line-graph formulation attains a higher APLS score than all original-graph readouts. This result suggests that the improvement cannot be attributed solely to a more effective edge-scoring function. Instead, by representing candidate edges as nodes, the line graph enables edge embeddings to be refined through interactions with neighboring candidate edges before classification. Such edge-level message passing is particularly beneficial in topology-sensitive scenarios, including occlusions and non-planar crossings. Furthermore, when the overpass/underpass prediction head is added, the performance gap between the best original-graph readout and the line-graph formulation widens, further highlighting the advantage of edge-centric reasoning.

\begin{table}[ht]
\caption{Comparison between line-graph reasoning and original-graph link-prediction readouts on the City-scale dataset. All variants use the same sparse candidate graph and visual feature setting.}
\label{tab:readout}
\footnotesize
\centering
\begin{tabular}{lccccc}
\toprule
Readout & $N_{\text{sampled}}$ & Prec.$\uparrow$ & Rec.$\uparrow$ & F1$\uparrow$ & APLS$\uparrow$ \\
\midrule
Original graph + concat MLP & 2 & 91.81 & 73.34 & 81.35 & 65.72 \\
Original graph + concat MLP & 4 & 91.05 & 73.60 & 81.17 & 66.57 \\
Original graph + dot product & 4 & 92.38 & 70.38 & 79.60 & 61.13 \\
Original graph + bilinear & 4 & 91.29 & 74.04 & 81.57 & 67.51 \\
Original graph + Hadamard MLP & 4 & 90.51 & 74.50 & 81.54 & 67.33 \\
\midrule
SAM-Road~\cite{hetang2024segment} & -- & 90.47 & 67.69 & 77.23 & 68.37 \\
\textbf{Ours: line graph} & 4 & 91.09 & 75.44 & 82.37 & 68.88 \\
\midrule
Original graph + bilinear w/ overpass & 4 & 93.15 & 74.31 & 82.44 & 67.73 \\
\textbf{Ours w/ overpass} & 4 & 92.75 & 76.64 & 83.77 & 70.40 \\
\bottomrule
\end{tabular}
\end{table}

\subsection{Analysis of Sparse Graph Construction}
\label{sec:app_sparse_graph}

The proposed framework uses a radius-based sparse Euclidean graph to construct candidate road connections. Compared with local-subgraph methods, this global sparse graph preserves longer-range candidate connectivity while avoiding the cost of dense all-pair inference, which is important for resolving occlusions and non-planar crossings.

\begin{table}[ht]
\caption{Comparison with K-NN graph construction and the local-subgraph baseline on the City-scale dataset. All variants use $N_{\mathrm{sampled}}=4$ and the overpass/underpass head.}
\label{tab:knn}
\small
\centering
\begin{tabular}{lcccc}
\toprule
Graph & Prec.$\uparrow$ & Rec.$\uparrow$ & F1$\uparrow$ & APLS$\uparrow$ \\
\midrule
8-NN w/ overpass & 94.05 & 70.70 & 80.42 & 61.15 \\
16-NN w/ overpass & 92.83 & 74.06 & 82.16 & 67.33 \\
SAM-Road local subgraph~\cite{hetang2024segment} w/ overpass & 92.67 & 67.52 & 77.89 & 66.82 \\
\midrule
\textbf{Ours w/ overpass} & 92.75 & 76.64 & 83.77 & 70.40 \\
\bottomrule
\end{tabular}
\end{table}

We compare the radius-based graph with K-nearest-neighbor graph constructions and a local-subgraph baseline on the City-scale dataset. As shown in \cref{tab:knn}, the radius-based sparse graph achieves the best F1 and APLS among 8-NN, 16-NN, and the SAM-Road local-subgraph variant. The 8-NN graph has high precision but low recall, suggesting that it misses valid longer-range connections. The 16-NN graph improves recall but introduces more candidate edges, leading to lower APLS than our radius-based graph. Overall, the radius-based graph provides a better trade-off between candidate-edge recall and graph sparsity.

The complexity of constructing the line graph is $O(M+\sum_{v\in V}\deg(v)^2)$, where $M$ is the number of candidate edges and $\deg(v)$ is the degree of vertex $v$ in the candidate graph. Thus, the practical overhead follows the sparsity of the underlying graph, with the empirical trend 8-NN $\lesssim$ radius-based $\approx$ local-subgraph $<$ 16-NN.

\bibliographystyle{splncs04}
\bibliography{main}